\documentclass[10pt,twocolumn,letterpaper]{article}

\usepackage{cvpr}              % To produce the CAMERA-READY version

\definecolor{cvprblue}{rgb}{0.21,0.49,0.74}
\usepackage[pagebackref,breaklinks,colorlinks,allcolors=cvprblue]{hyperref}

\hypersetup{
    colorlinks,
    linkcolor={red!50!black},
    citecolor={blue!50!black},
    urlcolor={blue!80!black}
}

\definecolor{forestGreen}{RGB}{34, 139, 34}
\definecolor{firebrick}{RGB}{178, 34, 34}

\newcommand{\dataset}[1]{\texttt{#1}}
\newcommand{\model}[1]{\texttt{#1}}
\newcommand{\method}[1]{\texttt{#1}}
\newcommand{\baseline}[1]{\texttt{#1}}

\newcommand{\negxmark}{\color{firebrick}{×}}
\newcommand{\poscheckmark}{\color{forestGreen}{\checkmark}}

\newcommand{\improv}[1]{\textcolor{forestGreen}{(+#1)}}
\newcommand{\worse}[1]{\textcolor{firebrick}{(-#1)}}

\newcommand{\myxmark}{\color{forestGreen}{×}}
\newcommand{\mycheckmark}{\color{firebrick}{\checkmark}}

\usepackage{amsthm}
\usepackage{amsmath}

\usepackage{placeins}
\usepackage{hyperref}
\usepackage[capitalize]{cleveref} %added this because ref to Proposition shows Theorem instead
\usepackage{url}
\usepackage{booktabs}
\usepackage{multirow}
\usepackage{amsmath, amsfonts}
\usepackage{overpic}
\usepackage{amssymb}
\usepackage{appendix}
\usepackage{todonotes}
\usepackage{leftindex}
\usepackage{algorithm}

\usepackage{algorithmic}
\usepackage{dsfont}
\usepackage{colortbl}
\usepackage{tabularx}
\usepackage[accsupp]{axessibility}  % Improves PDF readability for those with disabilities.

\newcommand\blfootnote[1]{%
  \begingroup
  \renewcommand\thefootnote{}\footnote{#1}%
  \addtocounter{footnote}{-1}%
  \endgroup
}

\def\paperID{16387} % *** Enter the Paper ID here
\def\confName{CVPR}
\def\confYear{2025}

\title{\begin{center}Task Singular Vectors:\end{center} Reducing Task Interference in Model Merging \vspace{-0.15cm}}
\author{
Antonio Andrea Gargiulo$^{\star}$\hfill Donato Crisostomi$^{\star}$ \hfill Maria Sofia Bucarelli \vspace{0.1cm} \\ 
Simone Scardapane \hfill Fabrizio Silvestri \hfill Emanuele Rodolà \vspace{0.25cm} \\
Sapienza University of Rome \vspace{0.1cm} \\
{\tt\small gargiulo.1769185@studenti.uniroma1.it} \; \;  
{\tt\small \{crisostomi, rodola\}@di.uniroma1.it}\\
{\tt\small simone.scardapane@uniroma1.it} \; \; 
{\tt\small \{bucarelli, fsilvestri\}@diag.uniroma1.it
}
}

\newcommand{\ntasks}{T}

\theoremstyle{plain}
\newtheorem{theorem}{Theorem}[section]
\newtheorem{proposition}[theorem]{Proposition}

\newtheorem{corollary}[theorem]{Corollary}
\theoremstyle{definition}

\theoremstyle{remark}

\begin{document}

\maketitle

\begin{abstract}
    \baseline{Task Arithmetic} has emerged as a simple yet effective method to merge models without additional training. However, by treating entire networks as flat parameter vectors, it overlooks key structural information and is susceptible to task interference. In this paper, we study task vectors at the layer level, focusing on task layer matrices and their singular value decomposition. In particular, we concentrate on the resulting singular vectors, which we refer to as \emph{Task Singular Vectors} (TSV). Recognizing that layer task matrices are often low-rank, we propose \method{TSV-Compress} (\method{TSV-C}), a simple procedure that compresses them to 10\% of their original size while retaining 99\% of accuracy. We further leverage this low-rank space to define a new measure of task interference based on the interaction of singular vectors from different tasks. Building on these findings, we introduce \method{TSV-Merge} (\method{TSV-M}), a novel model merging approach that combines compression with interference reduction, significantly outperforming existing methods.
\end{abstract}

\section{Introduction}
%!Tex root=../main.tex
%
% Broader intro / context

The widespread availability of pre-trained models and public repositories has driven the development of techniques for efficiently combining and reusing existing models. Among these techniques, model merging approaches enable the creation of multi-task models without additional training. \blfootnote{$\star$ denotes equal contribution.}
One popular approach, \baseline{Task Arithmetic} ({TA}) \citep{ilharco_editing_2023}, stands out for its simplicity and effectiveness. However, by treating entire networks as high-dimensional vectors, {TA} and subsequent works overlook crucial structural information \citep{yadav_ties-merging_2023, yu_language_2024, jin_dataless_2023, rame_diverse_2023, matena_merging_2022, daheim2024model, davari2025model}. This flattened view limits these approaches to coarse-grained measures like cosine similarity for assessing inter-task interactions. 

%However, this method inherits challenges inherent in multi-task learning: {\em weight interference}, where specific parameters conflict across tasks , and {\em task interference}, where even non-overlapping weights can interfere globally \citep{ortiz-jimenez_task_2023, wang_localizing_2024}. 

% Intro to our work
In this work, we instead focus on preserving and leveraging the natural structure of neural networks by examining weight differences at the layer level while retaining their matrix form wherever possible. 
Our approach begins by examining per-layer task \emph{matrices}, which represent weight changes for each task, through singular value decomposition (SVD). This decomposition yields singular vectors and singular values that capture the most significant directions of variation within each layer. We term these singular vectors \emph{Task Singular Vectors} (TSV), as they provide an interpretable basis for assessing task-specific contributions at each layer. 
\begin{figure}
    \centering
    \includegraphics[width=\columnwidth]{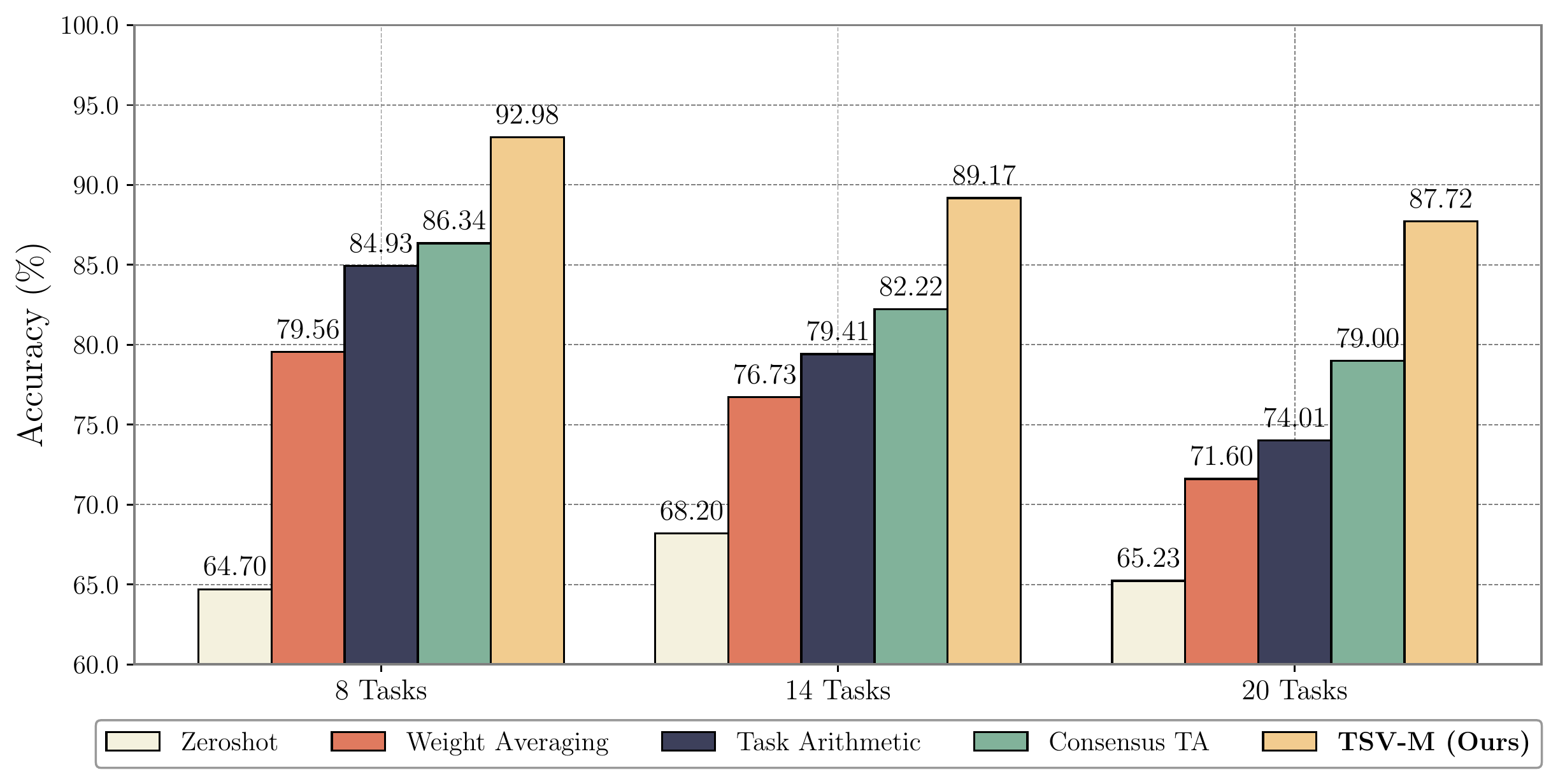}
    \caption{Mean accuracy of a \model{ViT-L-14} merged over 8, 14, and 20 tasks respectively. By significantly surpassing existing methods, \method{TSV-M} establishes the new state of the art in model merging.}
    \label{fig:main-results}
\end{figure}
%
% Layer task matrices are low-rank
%
Importantly, in accordance with recent literature on PEFT (cfr.~\cref{sec:related}), our analysis confirms that task matrices are inherently low-rank, meaning that only a small portion of TSVs is sufficient to represent the layer’s function with high fidelity.

% TSV-Compress
%
Building on this insight, we introduce \method{TSV-Compress} (\method{TSV-C}), a simple yet effective procedure to compress task vectors down to $10\%$ of their original size while maintaining $99\%$ of the original accuracy. Focusing on only a fraction of the most relevant TSVs for each task, \method{TSV-Compress} enables efficient storage and processing of task-specific information without waiving performance.

%
% Task interference measure
%
Beyond compression, examining the interplay of TSVs across different tasks provides a geometrically informed framework for analyzing task interference at the individual layer level. By assessing how singular vectors from different tasks align or diverge within each layer, this approach offers a significantly more fine-grained understanding of inter-task interactions, going beyond global vector similarity metrics like cosine similarity.
%
%
% Model merging
%
Building upon these principles, we introduce \method{TSV-Merge} (\method{TSV-M}), a novel model merging method that combines compression and task interference reduction. This is achieved by discarding the irrelevant singular vectors for each task and then reducing their inter-task interference with a whitening transformation over their similarity matrix. 
%
% Results
%
We empirically show our approach to effectively reduce task interference, and the reduction to be complementary to the compression step. We then evaluate our approach across several benchmarks and show it outperforms existing methods by an average of $\sim 15\%$ accuracy points, establishing the new state of the art by a large margin. 
We release all our code for research purposes\footnote{\href{https://github.com/AntoAndGar/task_singular_vectors}{https://github.com/AntoAndGar/task\_singular\_vectors}}.

%
% Contributions
%
In summary, our contribution is 4-fold:
\begin{itemize}
    \item We study the singular value decomposition of per-layer task matrices, showing them to be low-rank and measuring the interplay of singular vectors from different tasks;  
    \item We introduce \method{TSV-C}, a method that builds on this insight to compress task vectors by a factor of $10\times$ while preserving $99\%$ of their original accuracy;
    \item We propose \method{TSV-M}, a model merging technique that complements compression with a task interference reduction step by applying a whitening transformation to decorrelate singular vectors across tasks;
    \item We conduct extensive experiments on multiple computer vision datasets, showing that \method{TSV-M} significantly outperforms existing model merging methods and provide in-depth analyses to uncover the underlying factors contributing to its success.
\end{itemize}

\section{Related Work}\label{sec:related}

\begin{table}[htbp]
    \centering
    \small
    \caption{Model merging approaches and their requirements.}
    \label{tab:prerequisites}
    \begin{tabularx}{\linewidth}{@{}ccccc@{}}
        \toprule
        \multirow{2}{*}{\textbf{Method}} & \textbf{Additional} & \textbf{Extra} & \multicolumn{2}{c}{\textbf{Validat. data}} \\
        & \textbf{training} & \textbf{storage} & \textbf{inputs} & \textbf{labels} \\ 
        \midrule
        Weight Avg. \cite{wortsman_model_2022} & \myxmark & \myxmark & \myxmark & \myxmark \\
        \midrule
        Fisher-Merg. \cite{matena_merging_2022} & \myxmark & \myxmark & \mycheckmark & \myxmark \\
        RegMean \cite{jin_dataless_2023} & \myxmark & \myxmark & \mycheckmark & \myxmark \\
        \midrule
        EMR-Merging \cite{huang2024emrmerging} & \myxmark & \mycheckmark & \myxmark & \myxmark \\
        TALL-Mask \cite{wang_localizing_2024} & \myxmark & \mycheckmark & \myxmark & \myxmark \\
        \rowcolor{gray!15} \textbf{TSV-C (Ours)} & \myxmark & \mycheckmark & \myxmark & \myxmark \\
        \midrule
        Task Arith. \cite{ilharco_editing_2023} & \myxmark & \myxmark & \mycheckmark & \mycheckmark \\
        Ties-Merging \cite{yadav_ties-merging_2023} & \myxmark & \myxmark & \mycheckmark & \mycheckmark \\
        AdaMerging \cite{yang2024adamergingadaptivemodelmerging} & \mycheckmark & \myxmark & \mycheckmark & \myxmark \\
        Consensus-TA \cite{wang_localizing_2024} & \myxmark & \myxmark & \mycheckmark & \mycheckmark \\
        \rowcolor{gray!15} \textbf{TSV-M (Ours)} & \myxmark & \myxmark & \myxmark & \myxmark \\
        \bottomrule
    \end{tabularx}
\end{table}

\paragraph{Model Merging} offers an efficient alternative to ensembling by combining existing models without the need for further training. Several approaches address this by determining the neuron permutations that align the models into a shared optimization basin, after which the models are combined using a straightforward averaging method. \citep{ainsworth2023git, jordan_repair_2023, zip-it, rebasin-implicit-sinkhorn, cycle-consistent}. An alternative line of approaches focuses on the multi-task scenario, where a single pre-trained model is fine-tuned for different tasks \citep{ilharco_editing_2023, yadav_ties-merging_2023, yu_language_2024, matena_merging_2022, wortsman_model_2022, davari2025model, wang_localizing_2024, zhou2024atmimprovingmodelmerging,shen2024efficient}. The prerequisites for methods in this category are outlined in \Cref{tab:prerequisites}.
\baseline{Task Arithmetic} \citep{ilharco_editing_2023} introduces the concept of \emph{task vectors}, which are the weight differences between fine-tuned models and the pre-trained base model. By averaging these vectors, a merged multi-task model can be created; conversely, negating a task vector allows for forgetting a specific task.
\baseline{TIES} \citep{yadav_ties-merging_2023} addresses redundancies in model parameters by first selecting the top-$k$ most significant parameter changes and then constructing a sign vector based on the majority sign across all models. The latter is used to merge the task vectors disjointly, meaning the average is not computed when a parameter is zero or when parameters disagree in sign.
Similarly, \baseline{DARE} \citep{yu_language_2024} randomly resets redundant parameters to their pre-trained values and rescales the remaining parameters by a factor proportional to the dropped ones, aiming to reduce interference among tasks.
\baseline{Fisher Merging} \citep{matena_merging_2022} and \baseline{RegMean} \citep{jin_dataless_2023} merge models by performing weighted averaging, utilizing the Fisher information matrix and the inner product of input vectors, respectively.
\baseline{Model Breadcrumbs} \citep{davari2025model} focuses on merging only significant weights by discarding outliers and both minor and large perturbations in the fine-tuned parameters.
More recently, \citet{wang_localizing_2024} observed that tasks often utilize non-overlapping sets of weights. They propose two methods: the first, \baseline{TALL-Mask}, uses binary masks to activate important task-specific weights, requiring extra storage (as does our method \method{TSV-C}) for the masks. The second, \baseline{Consensus Merging}, leverages these masks to remove parameters that are important to less than two tasks.
Like our approach, \baseline{TwinMerging} \cite{lu2024twin} and \baseline{SMILE} \cite{tang2024smile} apply SVD to layer task matrices. However, neglecting singular vector interference, they require a router to selectively activate a subset during inference.

Unlike existing approaches, our methods explicitly address task interference by leveraging the geometric structure of singular vectors to minimize interactions between task-specific parameters. Rather than averaging or selectively merging parameters as prior techniques do, we decorrelate singular vectors to directly reduce interference across tasks. Additionally, we achieve targeted compression by isolating each task’s unique components, storing only essential parts, and ensuring minimal interference, a step beyond traditional SVD-based compression approaches.

\paragraph{SVD for Model Compression} A significant body of work explores low-rank decompositions for fully-connected and convolutional layers \citep{denil2013predicting, denton2014exploiting, jaderberg2014speeding, kolda2009tensor, garipov2016ultimate}, as well as tensor decompositions \citep{denton2014exploiting, kim2016compression, garipov2016ultimate}. These approaches typically achieve a layer-by-layer factorization of a trained network, focusing on minimizing the difference between the original and the low-rank approximated weight matrices. This is accomplished using techniques like SVD \citep{denton2014exploiting, garipov2016ultimate, jaderberg2014speeding, kolda2009tensor} or iterative optimization \citep{jaderberg2014speeding, kim2016compression}. While other methods such as weight quantization, knowledge distillation, and pruning exist, we utilize SVD not only for low-rank approximation and compression but also because its matrix decomposition properties make it particularly effective in analyzing and mitigating interference.

\section{Task Singular Vectors}
In this section, we introduce key concepts for understanding our approach, including a new measure of task interference that is central to our method.

\subsection{Background}\label{subsec:background}
We build on \baseline{Task Arithmetic} (TA), which defines task vectors capturing the differences in model weights for individual tasks. Formally, the weights $\theta_\text{MT}$ of a multi-task model for $\ntasks$ tasks are computed by aggregating the task-specific weight differences, or task vectors, as follows:
\begin{equation}
    \theta_\text{MT} = \theta_\text{pre} + \alpha \frac{\sum_{i=1}^\ntasks \tau_i}{\ntasks}\,, \label{eq:task-arithmetic}
\end{equation}
where $\theta_\text{pre}$ is the set of pretrained model weights, $\alpha$ is a scaling factor, and $\tau_i=\theta_{\text{ft}_i}-\theta_\text{pre}$ is the task vector for task $i$, with $\theta_{\text{ft}_i}$ being the fine-tuned weights for the task.
Differently from TA, however, we consider these operations at the layer level. From this perspective, \cref{eq:task-arithmetic} becomes
\begin{equation}
    \label{eq:merge_weights_layer}
    \theta_\text{MT}^{(l)} = \theta_\text{pre}^{(l)} + \alpha \frac{\sum_{i=1}^\ntasks \Delta_i^{(l)}}{\ntasks},
\end{equation}
where \( \theta_\text{pre}^{(l)} \) encodes the pretrained weights for layer \( l \), and $\Delta_i^{(l)} = \theta_{\text{ft}_i}^{(l)} - \theta_\text{pre}^{(l)} $
is the task-specific weight difference for task \( i \) at layer \( l \). When layer $l$ has a matrix structure, we call its corresponding $\Delta_i{^l}$ the \emph{per-layer task matrix} for task $i$. When this is not the case, our framework defaults to standard TA.
For brevity, we will generally omit the layer index and refer to the layer-$l$ task matrix $\Delta_i^l$ as $\Delta_i$. %Intuitively, task vectors can then be seen as the concatenation of flattened layer task matrices. 

\subsubsection*{Decomposing layer task matrices}
Treating layer-wise weights as structured entities rather than flattened vectors enables us to analyze their SVD, revealing low-rank properties and deeper insight into the inter-task interactions.
Given two tasks $i, j$, we consider the SVD of their task matrices $\Delta_i$ and $\Delta_j$ at a generic layer:
\begin{equation*}
    \Delta_i = U_i \Sigma_i V_i^\top, \qquad 
    \Delta_j = U_j \Sigma_j V_j^\top
\end{equation*}
where \(U_i, U_j\) and \(V_i, V_j\) are the matrices of left and right singular vectors respectively, and \(\Sigma_i,\Sigma_j\) are diagonal matrices of singular values. Due to their role in assessing task-specific contributions, we term the obtained singular vectors \emph{Task Singular Vectors} (TSV).

Importantly for our treatment, we can equivalently write the aggregated task matrices in \cref{eq:merge_weights_layer} in terms of their singular vectors and values. By defining $U=[U_1  \cdots  U_{\ntasks}]$ as the column-wise concatenation of all the left TSVs, $V = [V_1 \cdots V_{\ntasks}]$ as the row-wise concatenation of the right TSVs, and $\Sigma$ as the block diagonal matrix with $\{ \Sigma_i \}_{i=1}^\ntasks$ along its diagonal, we can write:
\begin{equation}\label{eq:M}
    M = U \Sigma V^\top = \sum_{i=1}^\ntasks U_i \Sigma_i V^{\top}_i = \sum_{i=1}^\ntasks \Delta_i \,.
\end{equation}
In the simple case where $T=2$, $M$ would be given by:
\begin{equation*}
    M = \left[
\begin{array}{c c}
    U_{1} & U_{2} \\
\end{array}
\right]
\left[
\begin{array}{c c}
    \Sigma_{1} &  0 \\
    %\hline
    0 & \Sigma_{2} \\
\end{array}
\right]
\left[
\begin{array}{c}
    V_{1}^\top \\
    %\hline
    V_{2}^\top \\
\end{array}
\right] = \Delta_1 + \Delta_2 \,.
\end{equation*}
When concatenating singular components from different tasks, we violate certain properties of the SVD. Specifically, the matrices \( U \) and \( V \) become non-orthogonal because singular vectors from different tasks may overlap. Additionally, the singular values \( \Sigma_i \) from different tasks can vary significantly in magnitude, which may bias the merging process toward tasks with larger singular values.
\subsection{Low-rank nature of layer task matrices} \label{subsec:low-rank}
\begin{figure}
  \centering
  \includegraphics[width=\linewidth]{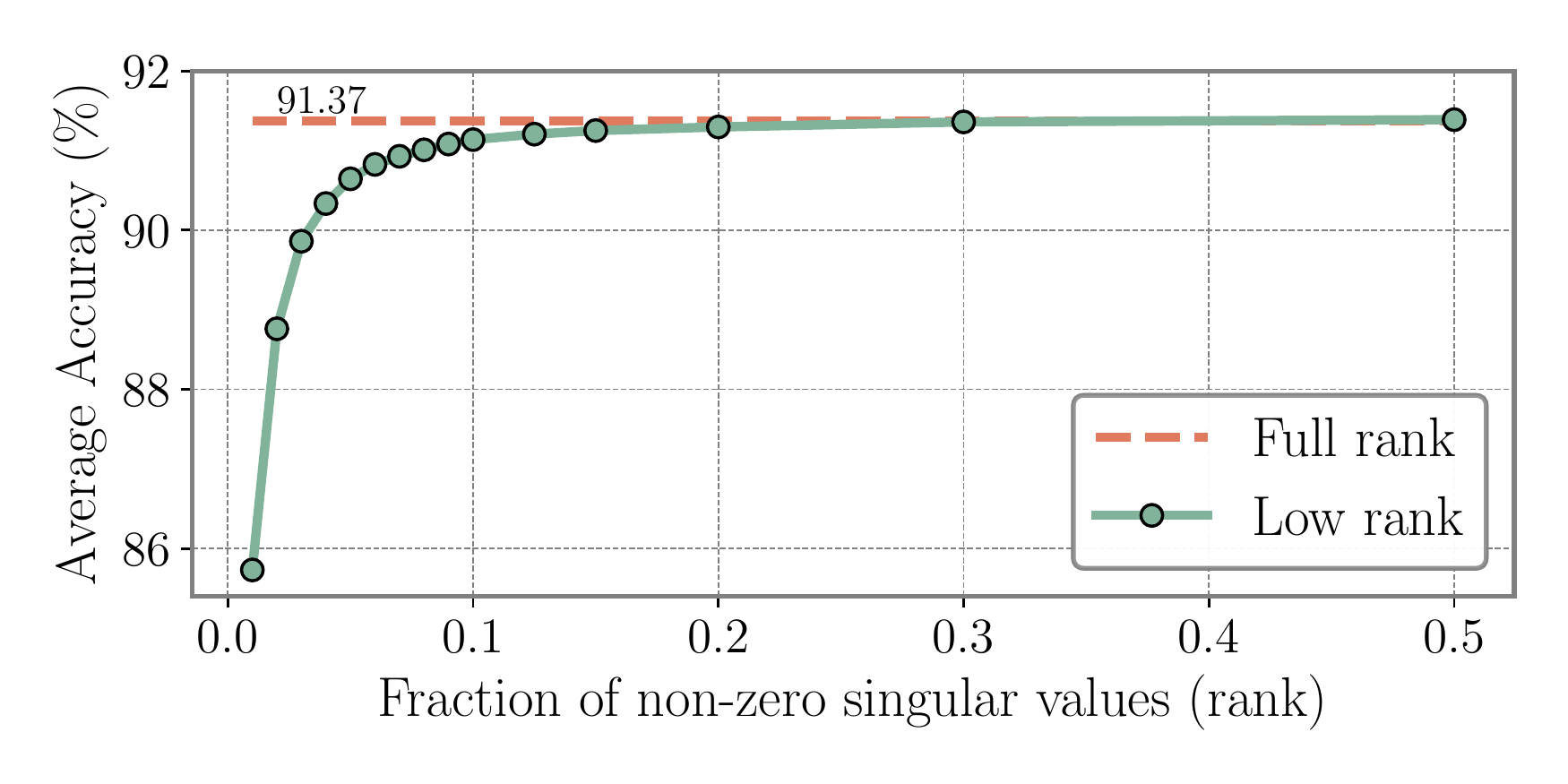}
    \caption{Mean absolute accuracy of the \model{ViT-B-32} model across increasing fractions of retained singular components, averaged over 20 tasks. The red line represents the average accuracy of the original fine-tuned models with full-rank task matrices, while the green line shows the accuracies using low-rank approximations.}
  \label{fig:acc_by_rank}
\end{figure}

We start our analysis by studying the low-rank properties of per-layer task matrices. Interpreting SVD as a sum of rank-one matrices, for each task $i$, by Eckart-Young's theorem \citep{Eckart1936TheAO} we get the best approximation (in Frobenius norm) of each task matrix $\Delta_i$ by retaining only the top-$k$ singular values and their corresponding vectors:
\begin{equation}
\hat{\Delta}_{i} = \sum_{j=1}^{k} \sigma_j^{i} u_j^{i} v_j^{i\top}.
\end{equation}
As we show in \cref{fig:acc_by_rank}, we find task matrices to be inherently low-rank: a small subset of TSVs is sufficient to represent the layer's function with high fidelity. In particular, even when preserving only 3\% of the singular components per task, the mean accuracy drops by merely 1.5\%. This is noticeable considering that 97\% of singular components in each layer matrix are discarded. Building on this insight, we propose in \cref{subsec:tsv-compress} a compression algorithm that maintains $99\%$ of the accuracy while shrinking the task vectors to $10\%$ of their original size.  

Given this low-rank structure, it is natural to aggregate task matrices within their subspaces. We, therefore, obtain a reduced version of the aggregation matrix $M$ (\cref{eq:M}) as:
\begin{equation}\label{eq:sum-of-rank-1-matrices-multi-task}
    \hat{M} = \sum_{i=1}^{\ntasks} \hat{U}_i \hat{\Sigma}_i \hat{V}_i^\top = \sum_{i=1}^{\ntasks} \sum_{j=1}^{k} \sigma_j^{i} u_j^{i} v_j^{i\top},
\end{equation}
where \( \hat{U}_i \) and \( \hat{V}_i \) contain the top-\( k \) left and right singular vectors for task \( i \), respectively, and \( \hat{\Sigma}_i \) is the diagonal matrix of the top-\( k \) singular values \( \sigma_j^{i} \). This formulation highlights that the low-rank approximation \( \hat{M} \) utilizes the most relevant singular components from each task to effectively combine the task-specific weight differences.
\begin{figure*}
  \centering
  \includegraphics[width=0.9\linewidth]{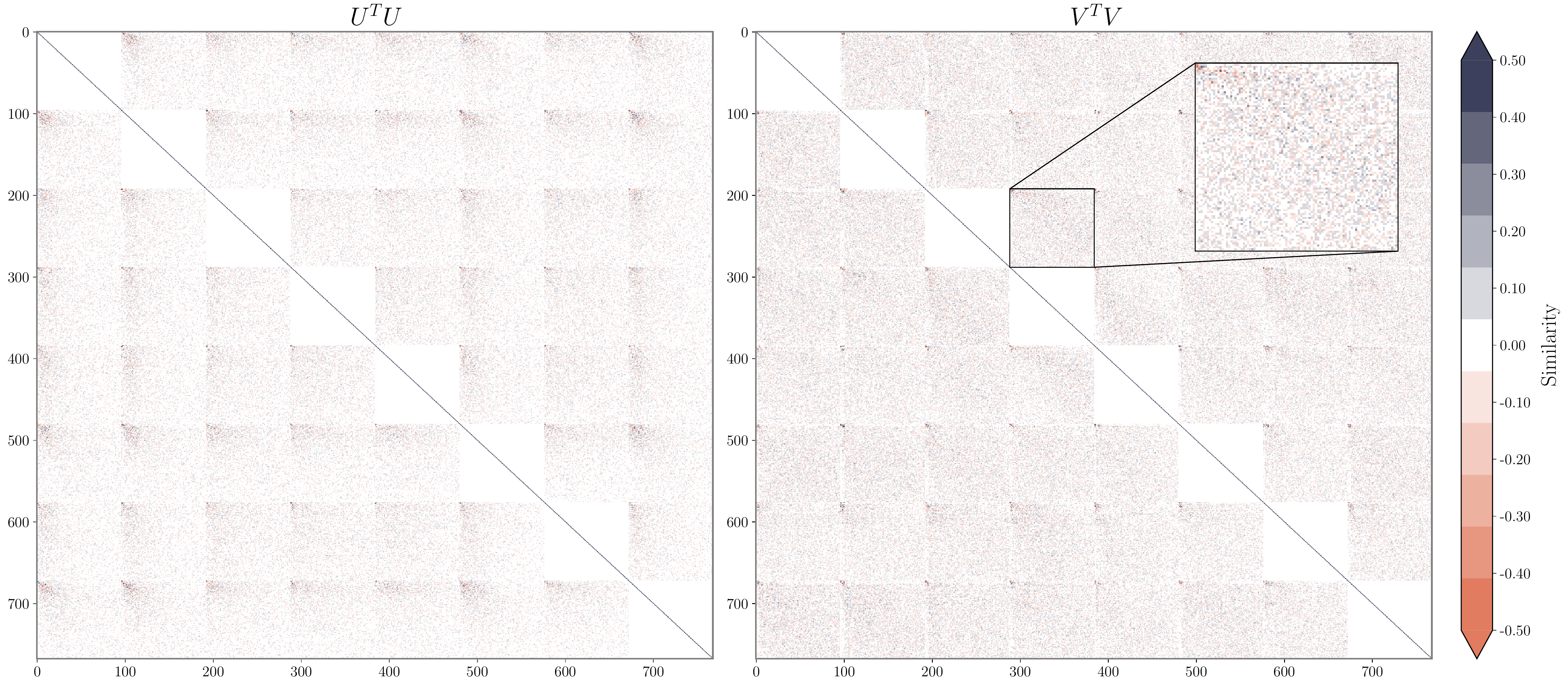}
  \caption{Visualization of task interference among 8 tasks computed on the first attention layer of a \model{ViT-B-32}. The diagonal blocks display intra-task similarities, while the off-diagonal blocks illustrate inter-task similarities. The zoomed-in section highlights the interaction between the right singular vectors of the 3$^{\text{rd}}$ and 4$^{\text{th}}$ tasks.}
  \label{fig:similarity_matrices}
\end{figure*}
\subsection{Singular Task interference}\label{subsec:task-interference}
\label{sec:STI}
%Quantifying task-specific contributions at each layer, Task Singular Vectors provide a geometrically informed tool to study task interference. 
We hereby introduce a score of task interference based on the interplay of TSVs from different tasks, which we term \emph{Singular Task Interference} (STI) 
\begin{equation}\label{eq:task-interference}
    \operatorname{STI}\left(\ \left\{\Delta_i\right\}_{i=1}^{T}\right) = \lVert (U^\top U - I) \Sigma (V^\top V - I) \rVert_1\,,
\end{equation}
where $U$, $\Sigma$ and $V$ are obtained by concatenating the singular value decompositions $(\left\{U_i\right\}_{i=1}^T$, $\left\{\Sigma_i\right\}_{i=1}^T$ $\left\{V_i\right\}_{i=1}^T)$ of per-layer task matrices $\left\{\Delta_i\right\}_{i=1}^{T}$ as detailed in \cref{subsec:background}.  
%
%examine the alignment of singular vectors in the corresponding task matrices through the inner products \( U^\top U \) and \( V^\top V \). 
In the expression above, high inner product values for \( U^\top U \) and \( V^\top V \) imply a higher likelihood of interference, with minimal interference ideally yielding identity matrices.

The underlying intuition is that overlapping singular vectors %-- whether partially similar or fully congruent --
suggest shared features in the weight space across tasks. Such overlap can introduce interference when models are merged, ultimately degrading performance on individual tasks.
We refer to \cref{fig:similarity_matrices} for a real example over eight tasks.

\section{Approach}
We demonstrate below how TSVs can be used for both compression and task interference reduction.

\begin{figure*}
  \centering
  \includegraphics[width=\linewidth]{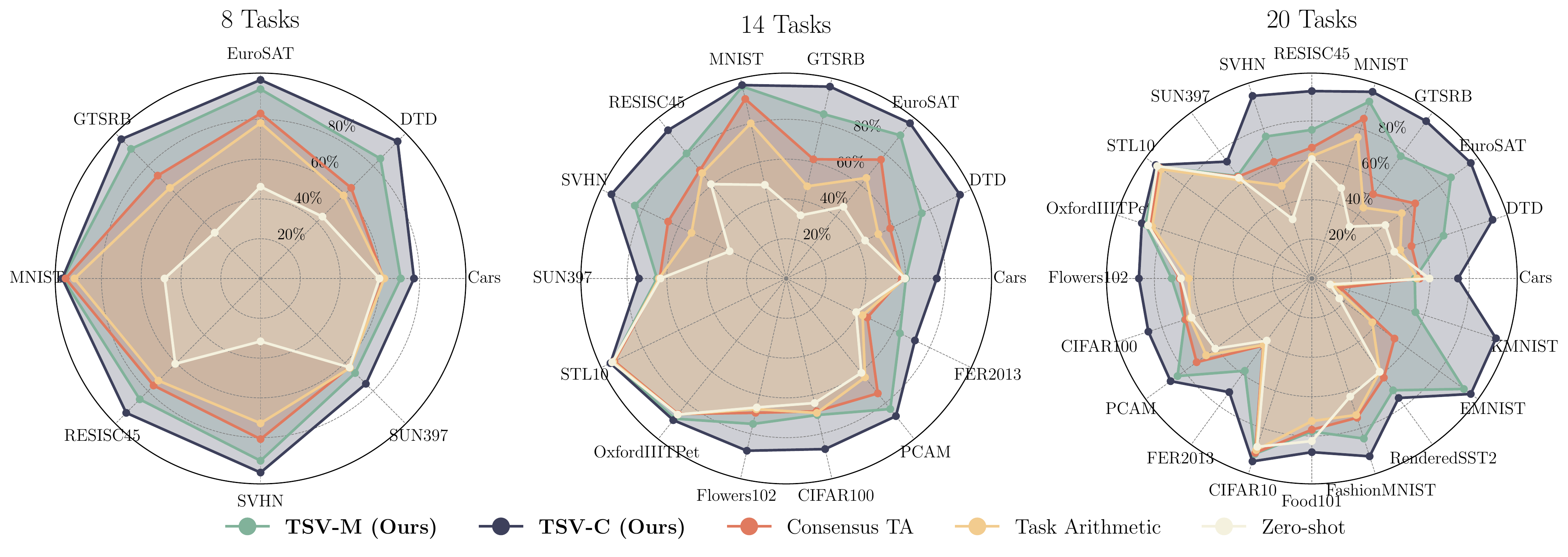}
  \caption{Absolute accuracy of a \model{ViT-B-32} merged over 8, 14, and 20 tasks, respectively.}
  \label{fig:radar_chart}
\end{figure*}

\subsection{TSV for compression}\label{subsec:tsv-compress}
%
%In certain scenarios, information regarding the task at hand is either available or inferable. In the first case, the optimal solution would be to use the task-specific fine-tuned model, boiling the problem down to model compression. In the second case, a router, such as those used in Mixture of Experts (MoE) techniques, can be used to infer the task and subsequently employ the corresponding task-specific model.
When the task is known or inferred (e.g. with a router, as in Mixture-of-Experts \cite{eigen2014learningfactoredrepresentationsdeep, shazeer2017} techniques), we can leverage the low-rank structure of per-layer task matrices (see \cref{subsec:low-rank}) to effectively compress these ones while discarding task singular vectors from different tasks.
%
%Capitalizing on the low-rank structure of per-layer task matrices seen in \cref{subsec:low-rank}, we hereby propose a simple compression technique to deal with such scenarios.
In particular, given a known or inferred task index $h$, we set other task components to zero, reducing \cref{eq:sum-of-rank-1-matrices-multi-task} to:
\begin{equation}
    \hat M = \sum_{i=1}^\ntasks \underset{[i=h]}{\mathds{1}} \sum_{j=1}^{k} \sigma_j^{i} u_j^{i} v_j^{i\top} = \sum_{j=1}^{k} \sigma_j^{h} u_j^{h} v_j^{h\top} = \hat{\Delta}_{i} \,.
\end{equation}
This formula yields a low-rank approximation of $\Delta_h$, where only the top $k$ singular components of the task-specific matrix are retained. Here, the same number of components $k$ is taken from each task to ensure that each layer is compressed by a factor of $\frac{1}{T}$ relative to the original dimension. Increasing $k$ improves approximation but reduces the compression.

\subsection{TSV for model merging} \label{subsec:tsv-merge}
In scenarios where the task identity is not known beforehand, we address the standard model merging problem by combining Task Singular Vectors (TSVs) to create a single model that performs well jointly across all tasks.

To reduce our measure of task interference, we decorrelate the TSVs of different tasks (encoded as columns in $\hat{U}$ and $\hat{V}$) by whitening these matrices to minimize their correlations. This can be achieved by applying the transformation $X \mapsto X(X^\top X)^{-\frac{1}{2}}$ to both $\hat{U}$ and $\hat{V}$.
For improved numerical stability, we reformulate this whitening as an orthogonal Procrustes problem, seeking the orthogonal matrix $\hat{U}_\bot$ that minimizes the projection error:
\begin{equation}
    \min_{\hat{U}_\bot} \| \hat{U}_\bot - \hat{U} \|_F \quad \text{s.t.}~\hat{U}_\bot^\top \hat{U}_\bot = I \,,
\end{equation}
and similarly for $\hat{V}$. This problem admits a closed-form solution via the SVD $\hat{U}=PDQ^\top$, yielding $\hat{U}_\bot = P Q^\top$ (similarly for $\hat{V}_\bot$).
\begin{proposition} \label{prop:whitening-procrustes}
    The transformations $X \mapsto  X(X^\top X)^{-\frac{1}{2}}$ (whitening) and $X \mapsto PQ^\top$ (Procrustes), where $X = PDQ^\top$ is the SVD of $X$, are equivalent.
\end{proposition}
\begin{proof}
    Given $X=PDQ^\top$ and recalling that $D^\top D=D^2$ for diagonal matrices, simple algebraic manipulation yields $X^\top X = Q D^2 Q^\top$. It follows that $(X^\top X)^{-1/2}=Q D^{-1} Q^\top$, therefore the whitening transformation can be rewritten as $X (X^\top X)^{-1/2}=(PDQ^\top)(Q D^{-1} Q^\top)=PDID^{-1}Q^\top=PQ^\top$, completing the proof.
\end{proof}
Once the (rank-reduced) TSV matrices $\hat{U}$ and $\hat{V}$ are decorrelated, we reconstruct the merged layer as in \cref{eq:sum-of-rank-1-matrices-multi-task}; see \cref{alg:tsvmerge} for the complete steps.

%$U^\top U  = V \Sigma^2 V^T ... V \Sigma^{-1} V^T$

%$U V^T = U\Sigma V^\top (V \Sigma^{-1} V^\top)$

\begin{algorithm}
    \caption{\method{TSV-Merge}.}\label{alg:tsvmerge}
    \begin{algorithmic}[1]
        \REQUIRE Task matrices $\Delta_1, \dots, \Delta_T$, scaling factor $\alpha$
        \ENSURE Merged model weights $\theta_{\text{MT}}$
        \FOR{$i = 1$ to $T$} 
            \STATE Compute SVD: $\Delta_i = U_i \Sigma_i V_i^\top$
            \STATE Retain first $\frac{1}{T}$ singular components of $U_i$, $\Sigma_i$, and $V_i$
        \ENDFOR
        \STATE \textbf{Concatenate} the matrices:
        \STATE \quad $U \gets [U_1\,|\,U_2\,|\,\dots\,|\,U_T]$
        \STATE \quad $\Sigma \gets \text{block-diag}(\Sigma_1, \Sigma_2, \dots, \Sigma_T)$
        \STATE \quad $V \gets [V_1\,|\,V_2\,|\,\dots\,|\,V_T]$
        \STATE \textbf{Compute} the SVD of $U$ and $V$:
        \STATE \quad $U = P_U D_U Q_U^\top$ \qquad $V = P_V D_V Q_V^\top$
        \STATE \textbf{Obtain} the orthogonal matrices:
        \STATE \quad $U_\bot = P_U Q_U^\top$ \qquad $V_\bot = P_V Q_V^\top$
        \STATE \textbf{Reconstruct} the merged matrix:
        \STATE \quad $\hat{M} \gets U_\bot \Sigma V_\bot^\top$
        \STATE \textbf{Construct} merged model weights:
        \STATE \quad $\theta_{\text{MT}} \gets \theta_{\text{pre}} + \alpha \hat{M}$
        \STATE \textbf{return} $\theta_{\text{MT}}$
    \end{algorithmic}
\end{algorithm}

\section{Results}
We evaluate our approaches over three different suites of tasks having cardinality 8, 14, and 20, respectively. The first one, introduced in \citep{ilharco_editing_2023}, consists of datasets: \dataset{Cars} \cite{krause_3d_2013}, \dataset{DTD} \cite{cimpoi_describing_2014}, \dataset{EuroSAT} \cite{helber_eurosat_2019}, \dataset{GTSRB} \cite{stallkamp_german_2011}, \dataset{MNIST} \cite{726791}, \dataset{RESISC45} \cite{cheng_remote_2017}, \dataset{SUN397} \cite{xiao_sun_2016}, and \dataset{SVHN} \cite{netzer_reading_nodate}.
The benchmark with 14 tasks builds on the preceding one, incorporating six additional datasets: \dataset{CIFAR100} \cite{krizhevsky_learning_nodate}, \dataset{STL10} \cite{coates_analysis_2011}, \dataset{Flowers102} \cite{nilsback_automated_2008}, \dataset{OxfordIIITPet} \cite{parkhi_cats_2012}, \dataset{PCAM} \cite{veeling_rotation_2018}, and \dataset{FER2013} \cite{goodfellow_challenges_2013}.
Finally, the 20-tasks benchmark includes the preceding 14 plus the following six: \dataset{EMNIST} \cite{cohen_emnist_2017}, \dataset{CIFAR10} \cite{krizhevsky_learning_nodate}, \dataset{Food101} \cite{bossard_food-101_2014}, \dataset{FashionMNIST} \cite{xiao_fashion-mnist_2017}, \dataset{RenderedSST2} \cite{socher_recursive_nodate}, and \dataset{KMNIST} \cite{clanuwat_deep_2018}.

% tabella con I checkpoints di tall mask
\begin{table*}
  \resizebox{1\textwidth}{!}{
    \begin{tabular}{cccc|ccc|ccc}
      \toprule
      \multirow{2}{*}{\textbf{Method}}         & \multicolumn{3}{c}{\model{ViT-B-32}}                     & \multicolumn{3}{c}{\model{ViT-B-16}}                     & \multicolumn{3}{c}{\model{ViT-L-14}}                    \\ 
                                               \cmidrule{2-10}
                                               & 8 tasks           & 14 tasks          & 20 tasks           & 8 tasks           & 14 tasks          & 20 tasks          & 8 tasks           & 14 tasks          & 20 tasks          \\ 
                                               \midrule
      \multicolumn{1}{c|}{Zeroshot}            & $48.26_{(53.59)}$ & $57.21_{(63.69)}$ & $56.10_{(62.41)}$ & $55.34_{(59.34)}$ & $61.28_{(66.19)}$ & $59.73_{(64.52)}$ & $64.70_{(68.00)}$ & $68.20_{(72.15)}$ & $65.23_{(68.99)}$ \\
      \multicolumn{1}{c|}{Weight Averaging}    & $66.34_{(72.13)}$ & $64.34_{(71.12)}$ & $61.04_{(67.53)}$ & $72.22_{(76.60)}$ & $69.46_{(74.82)}$ & $65.31_{(70.36)}$ & $79.56_{(83.15)}$ & $76.73_{(81.10)}$ & $71.60_{(75.60)}$ \\
      \multicolumn{1}{c|}{Task Arithmetic}     & $70.79_{(76.55)}$ & $65.32_{(72.09)}$ & $60.52_{(66.79)}$ & $75.41_{(79.58)}$ & $70.52_{(75.89)}$ & $65.78_{(70.76)}$ & $84.93_{(88.65)}$ & $79.41_{(83.95)}$ & $74.01_{(78.07)}$ \\
      \multicolumn{1}{c|}{Consensus TA}        & $75.03_{(80.84)}$ & $70.39_{(77.36)}$ & $65.43_{(71.98)}$ & $79.39_{(83.86)}$ & $74.39_{(79.92)}$ & $69.76_{(74.93)}$ & $86.34_{(90.08)}$ & $82.22_{(86.94)}$ & $79.00_{(83.22)}$ \\
      \rowcolor{gray!15}
      \multicolumn{1}{c|}{\textbf{\texttt{TSV-M} (Ours)}} & $\mathbf{85.86_{(92.31)}}$ & $\mathbf{80.06_{(87.88)}}$ & $\mathbf{77.07_{(84.29)}}$ & $\mathbf{89.01_{(93.94)}}$ & $\mathbf{84.58_{(91.01)}}$ & $\mathbf{80.57_{(86.45)}}$ & $\mathbf{92.98_{(96.98)}}$ & $\mathbf{89.17_{(94.43)}}$ & $\mathbf{87.72_{(92.50)}}$ \\
      \bottomrule
    \end{tabular}
  }
  \caption{Average absolute accuracy results on model merging benchmarks; subscript (in parentheses) is the normalized average accuracy.}
  \label{tab:task_acc}
\end{table*}

\subsection{Model merging results} \label{sec:MM_results}
We evaluate our method on three variants of the \baseline{CLIP} \cite{radford2021learningtransferablevisualmodels} model, each employing a different size of \model{ViT} \cite{dosovitskiy2021imageworth16x16words} visual encoder: \model{ViT-B-32}, \model{ViT-B-16}, and \model{ViT-L-14}. The main benchmarks involve merging 8, 14, and 20 tasks, mirroring the experimental setup described in \citet{wang_localizing_2024}. We compare our approach against several training-free model merging methods, including weight averaging, \baseline{Task Arithmetic} \citep{ilharco_editing_2023}, and \baseline{Consensus Merging} \citep{wang_localizing_2024}. For reference, we include the performance of zero-shot models in \Cref{tab:task_acc} to represent the minimum achievable accuracy, and the average of individually fine-tuned models in \Cref{tab:task_acc_compression} to indicate the maximum potential gains. Performance metrics were assessed using both the average absolute accuracy and the average normalized accuracy, calculated as detailed in \Cref{sec:normalized_accuracy}.

As presented in \Cref{tab:task_acc}, our method achieves state-of-the-art results across all benchmarks, regardless of the \model{ViT} size or the number of tasks involved. Notably, the most significant improvements were observed with the smaller \model{ViT-B-32} encoder, where our approach outperforms \baseline{Task Arithmetic} and \baseline{Consensus TA} by an average of $+15.45\%$ and $+10.71\%$ absolute accuracy, respectively. Furthermore, when utilizing the \model{ViT-L-14} model, our method attains an average normalized accuracy of $96.98\%$. This indicates that we can effectively replace eight individual task-specific models with a single merged model, incurring only a $-3.02\%$ reduction in average accuracy. These results highlight the promise of our model merging technique as a cost-efficient alternative to mixtures of experts, multi-task learning, and ensembling methods. We report in \cref{fig:radar_chart} the per-dataset accuracies.

% TSV-C VS TALL-masks
% tall mask checkpoints
\begin{table*}
  \resizebox{1\textwidth}{!}{
    \begin{tabular}{cccc|ccc|ccc}
      \toprule
      \multirow{2}{*}{\textbf{Method}}               & \multicolumn{3}{c}{\model{ViT-B-32}} & \multicolumn{3}{c}{\model{ViT-B-16}} & \multicolumn{3}{c}{\model{ViT-L-14}}                                                                                                                           \\ \cmidrule{2-10}
                                                     & 8 tasks                               & 14 tasks                              & 20 tasks                               & 8 tasks            & 14 tasks          & 20 tasks          & 8 tasks            & 14 tasks          & 20 tasks          \\ \midrule
      \multicolumn{1}{c|}{Finetuned}                 & $92.83_{(100)}$                       & $90.88_{(100)}$                       & $91.37_{(100)}$                       & $94.64_{(100)}$    & $92.76_{(100)}$   & $93.17_{(100)}$   & $95.81_{(100)}$    & $94.29_{(100)}$   & $94.73_{(100)}$   \\
      \multicolumn{1}{c|}{TALL-Mask+TIES}            & $93.13_{(100.37)}$                    & $90.92_{(100.04)}$                    & $91.11_{(99.70)}$                     & $94.68_{(100.04)}$ & $92.69_{(99.90)}$ & $93.05_{(99.87)}$ & $95.96_{(100.15)}$ & $93.40_{(99.09)}$ & $93.91_{(99.16)}$ \\
      \multicolumn{1}{c|}{\textbf{\texttt{TSV-C} (Ours)}} & $92.62_{(99.74)}$                     & $90.29_{(99.28)}$                     & $90.64_{(99.14)}$                     & $94.47_{(99.79)}$  & $92.25_{(99.41)}$ & $92.53_{(99.27)}$ & $95.68_{(99.85)}$  & $94.04_{(99.72)}$ & $94.42_{(99.66)}$ \\
      \bottomrule
    \end{tabular}
  }
  \caption{Average absolute accuracy results across all compression benchmarks for different models and varying number of tasks, subscript (in parentheses) the normalized average accuracies. Our TSV-C compression method consistently achieves over 99\% of the original fine-tuned models' accuracy while significantly reducing storage requirements.}
  \label{tab:task_acc_compression}
\end{table*}
\subsection{Compression results} \label{subsec:compression-results}
In \Cref{tab:task_acc_compression} we report the results for our compression algorithm \method{TSV-C}. We compare our method with \baseline{TALL-masks} \citep{wang_localizing_2024}, which stores very sparse binary masks for each task. 
As shown in \Cref{tab:task_acc_compression}, our method always retains more than $99\%$ of the original accuracy for all considered benchmarks and models. The results of \method{TSV-C} are comparable to those of \baseline{TALL-masks} \citep{wang_localizing_2024}, with the former slightly underperforming \baseline{TALL-masks} for small \model{ViTs} and the latter gaining the upper hand in the larger-scale one. In general, however, the two approaches differ by less than $1\%$ accuracy on average.
Regarding the storage, our method stores only the top $\frac{1}{T}$ singular vectors per task, which leads to a fixed storage requirement of approximately twice the size of the original model, regardless of the number of tasks. By storing binary masks, \baseline{TALL-masks} instead results in a storage size that varies with their compressibility, increasing with the number of tasks. This variability can lead to unpredictable storage requirements and may diminish compression benefits when masks are less compressible.

Our results show that \method{TSV-C} effectively balances compression and performance by leveraging the most significant TSVs. Maintaining near-original accuracy, our approach is particularly suitable for scenarios where storage constraints are critical but high model performance is still required.

%%%%%%%%%%%%%%%%%%%%%%%%%%%%%%%%%%%%%%%
% IMPORTANT: ACCURACY WITH GROWING NUMBER OF TASKS
%%%%%%%%%%%%%%%%%%%%%%%%%%%%%%%%%%%%%%%
% \textcolor{red}{As the number of tasks grows, the limited number of singular vectors in our method might not suffice to approximate each task's transformation accurately. However, as shown in Figure \ref{fig:acc_by_rank}, even when preserving only 3\% of the singular components per task, the mean accuracy drops by merely 1.5\%. This is noticeable considering that 97\% of singular components in each layer matrix are discarded. This indicates that our compression algorithm maintains high performance despite significantly reducing storage usage.}

% In our compression scheme, the storage requirement remains bounded at approximately twice the size of the original model for up to 33 tasks. Beyond this point, one can choose to accept a slight performance degradation to maintain the storage constraint or allow the storage requirement to scale linearly with the number of tasks, proportional to the percentage of singular components preserved.     

\section{Analysis}
%!Tex root=../main.tex
%
%%%%%%%%%%%%%%%%%%%%%%%%%%%%%%%%%%%%%%%%%%%%%%
% Intro to analysis
%%%%%%%%%%%%%%%%%%%%%%%%%%%%%%%%%%%%%%%%%%%%%%
%
In this section, we begin by conducting an ablation study to assess the contributions of interference reduction and low-rank compression to the overall performance. Next, we study how task interference varies across layers of different depths. Finally, we empirically show that our method does not require tuning a scaling coefficient.
%
%%%%%%%%%%%%%%%%%%%%%%%%%%%%%%%%%%%%%%%%%%%%%%
% Ablation study
%%%%%%%%%%%%%%%%%%%%%%%%%%%%%%%%%%%%%%%%%%%%%%
%
\begin{table}
  \centering
  \resizebox{1\columnwidth}{!}{
  \begin{tabular}{ccccc}
    \toprule
    \vspace{-0.1cm}Low-rank       & Interf.  & \multicolumn{3}{c}{\model{ViT-B-32}}                                           \\
    \cmidrule{3-5}
    approx. & reduction           & 8 tasks                                  & 14 tasks                & 20 tasks                \\
    \midrule
    \negxmark     & \negxmark     & 76.5 (+0.0)                        & 72.1 (+0.0)       & 66.8 (+0.0)       \\
    \poscheckmark & \negxmark     & 75.2 \worse{1.3}                    & 71.0 \worse{1.1}   & 66.3 \worse{0.5}   \\
    \negxmark     & \poscheckmark & 82.6 \improv{7.4}                   & 75.7 \improv{4.7}  & 69.9 \improv{3.6}  \\
    \poscheckmark & \poscheckmark & 92.3 \improv{9.7}                   & 87.9 \improv{12.2} & 84.3 \improv{14.4} \\
    \bottomrule
  \end{tabular}
  }
  \caption{Comparison of different versions of \baseline{Task Arithmetic}, comprising either the low-rank approximation step, the interference reduction step, or both. The method performing both corresponds to the proposed \method{TSV-Merge}.}
  \label{tab:ablation-study}
\end{table}
\begin{figure}
    \centering
    \includegraphics[width=\columnwidth]{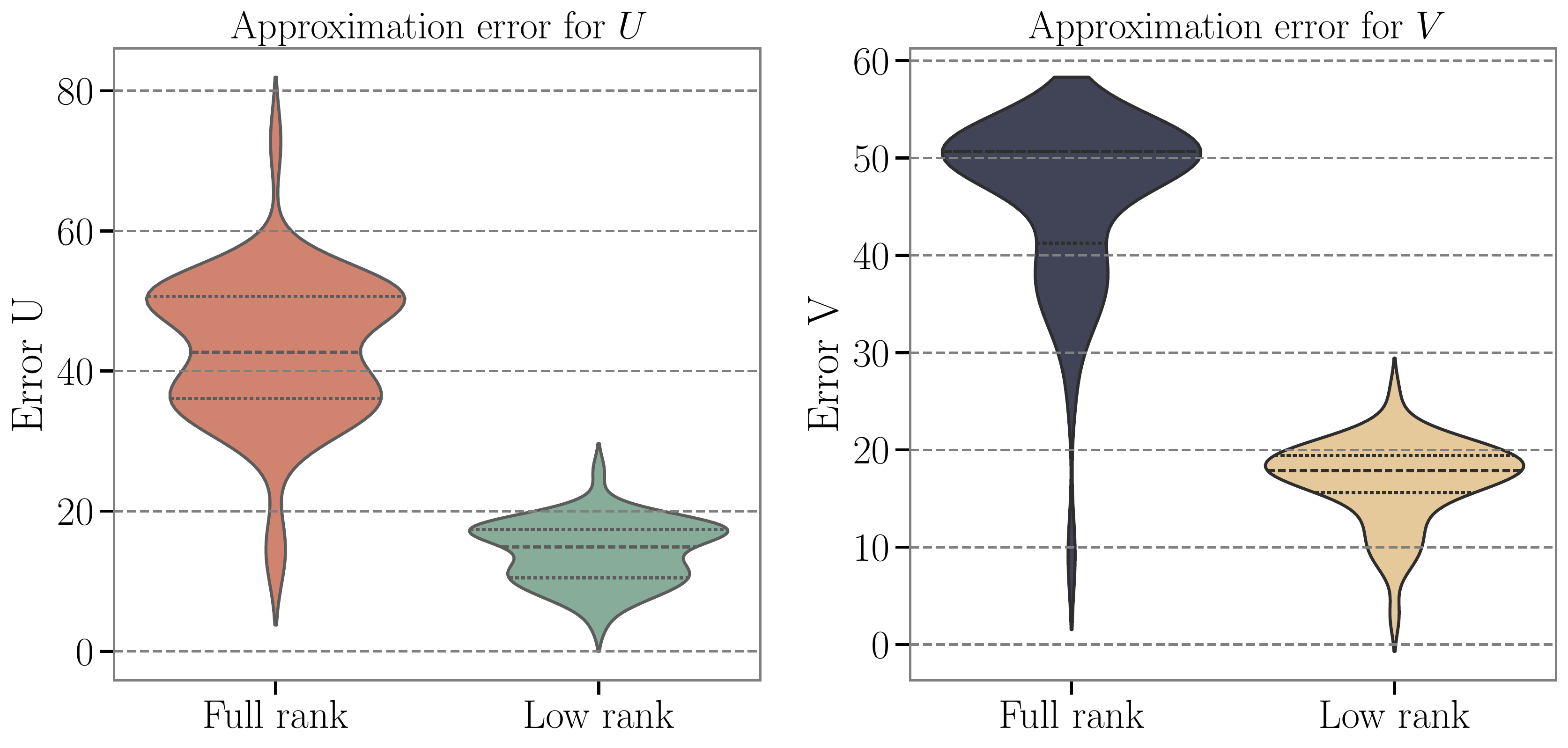}
    \caption{Approximation error from the orthogonalization of the TSVs through Procrustes for the \model{ViT-B-32} model across 8 tasks. The violin plots represent layer-wise approximation error distributions for $U$ and  $V$ in both full-rank and low-rank cases.
    %The approximation of the low rank task singular vectors shows a lower and more concentrated error.
    }
    \label{fig:approx-error}
\end{figure}
\subsection{Ablation study}
\label{subsec:ablation}
\method{TSV-M} combines low-rank approximation with task interference reduction. To evaluate the contribution of each component, we conducted an ablation study summarized in \Cref{tab:ablation-study}. Applying low-rank approximation alone to the layer task matrices results in worse performance than vanilla \baseline{Task Arithmetic}, shown in the first row where neither component is applied. In contrast, applying task interference reduction alone, while keeping full-rank matrices, significantly improves performance, with gains ranging from $+3.1\%$ to $+6.1\%$. However, the best results are achieved only when both steps are combined, yielding substantial performance improvements of $+15.8\%$ to $+17.5\%$. 

To explain why interference reduction applied to low-rank approximations in \method{TSV-M} outperforms its application to full-rank matrices, we analyze the errors introduced during the orthogonalization of the full-rank \( U \) and \( V \) matrices. As shown in \cref{fig:approx-error}, orthogonalizing full-rank matrices incurs significant approximation errors, measured by the sum of the Frobenius norms of reconstruction errors for the concatenated \( U \) and \( V \) matrices across all layers.

For example, in the \model{ViT-B-32} model with $8$ tasks, the full-rank setting exhibits a wider error distribution with a higher average, indicating greater variability and larger approximation discrepancies. In contrast, the low-rank setting produces a more compact and lower error distribution, suggesting better consistency in approximation across layers. In the following, we prove that this is not specific to the chosen model but a general property of our approach. The proof can be found in \Cref{mini_proof}. 
\begin{theorem}
\label{theo:error}
Let \(T \in \mathbb{N}\) such that \(T > 4\). Define \( U = [U_1, \dots, U_T] \) as the matrix obtained by concatenating \(T\) orthogonal matrices \( U_i \), each of shape \(n \times n\). Let \( \widehat{U} = [\widehat{U}_1, \dots, \widehat{U}_T] \) be the matrix formed by truncating each \( U_i \) to its first \(k\) columns. Denote by \(X\) and \(\widehat{X}\) the matrices resulting from Procrustes orthonormalization of \(U\) and \(\widehat{U}\), respectively. If \(k \leq n \frac{T - 2\sqrt{T}}{T}\), then 
    \[
    \| U - X \|_F \geq \| \widehat{U} - \widehat{X} \|_F \,.
    \]
\end{theorem}
Intuitively, the theorem asserts that the approximation error introduced by orthogonalization via Procrustes is smaller when applied to the concatenation of truncated matrices, compared to the concatenation of the original full-rank matrices, provided the rank of the truncated matrices satisfies certain conditions.
To better understand the statement of the theorem, consider the case \(T = 10\), as in our scenario. The term \(\frac{T - 2\sqrt{T}}{T} = \frac{10 - 2\sqrt{10}}{10} \geq \frac{1}{3}\). In our case, we choose \(k \approx \frac{n}{T}\). For \(T = 10\), this gives \(k \approx \frac{n}{10}\), which is indeed smaller than \(\frac{n}{3}\), satisfying the condition \(k \leq \frac{1}{3}n\).

%Notice that in our case, the condition \(  k \leq \frac{n}{T} \) is  satisfied.
%
We emphasize that other orthogonalization methods, such as Gram-Schmidt, are ineffective in this context because they do not preserve the overall structure of the original data. Gram-Schmidt sequentially orthogonalizes vectors without minimizing deviation from the original set, potentially resulting in significant distortions.

Finally, the substantial reduction in approximation error suggests that, while the advantages of interference reduction are considerable, they may be offset by the errors introduced when applied to full-rank matrices.
In contrast, starting with low-rank approximations captures the essential information of each task, making the orthogonalization step less costly in terms of approximation error. 

%%%%%%%%%%%%%%%%%%%%%%%%%%%%%%%%%%%%%%%%%%%%%
% PER-LAYER TASK INTERFERENCE
%%%%%%%%%%%%%%%%%%%%%%%%%%%%%%%%%%%%%%%%%%%%%

\begin{figure}
  \centering
  \includegraphics[width=\linewidth]{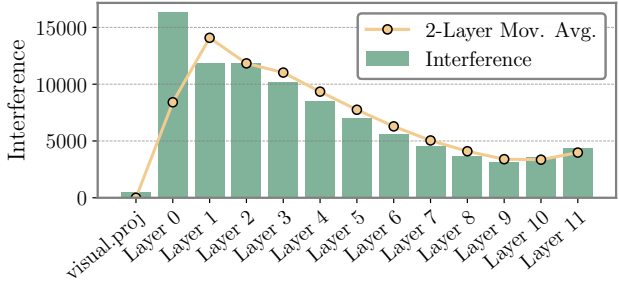}
  \caption{Singular Task Interference (STI) across layers in a \model{ViT-B-32} for 20 tasks. STI is high in early layers sharing common knowledge and lower in more specialized ones.}
  \label{fig:interference_by_layer}
\end{figure}
\subsection{Per-layer task interference}
We analyze task interference on a per-layer basis using \cref{eq:task-interference}. As shown in \cref{fig:interference_by_layer}, interference is highest in the initial transformer layers and decreases significantly in the deeper layers. This aligns with the understanding that early, generalized layers capture common features across tasks, increasing the potential for conflict, while deeper layers are more specialized for specific tasks, reducing shared representation and interference. 
For brevity, we grouped layers within each transformer block in our analysis; each ``Layer $n$'' in \cref{fig:interference_by_layer} includes two attention matrices and two MLP matrices. A layer-by-layer analysis is provided in \cref{fig:detailed_interference_by_layer}.

\subsection{Choice of the interpolation coefficient}
\label{sec:alpha}
\begin{figure}
  \centering
  \includegraphics[width=\linewidth]{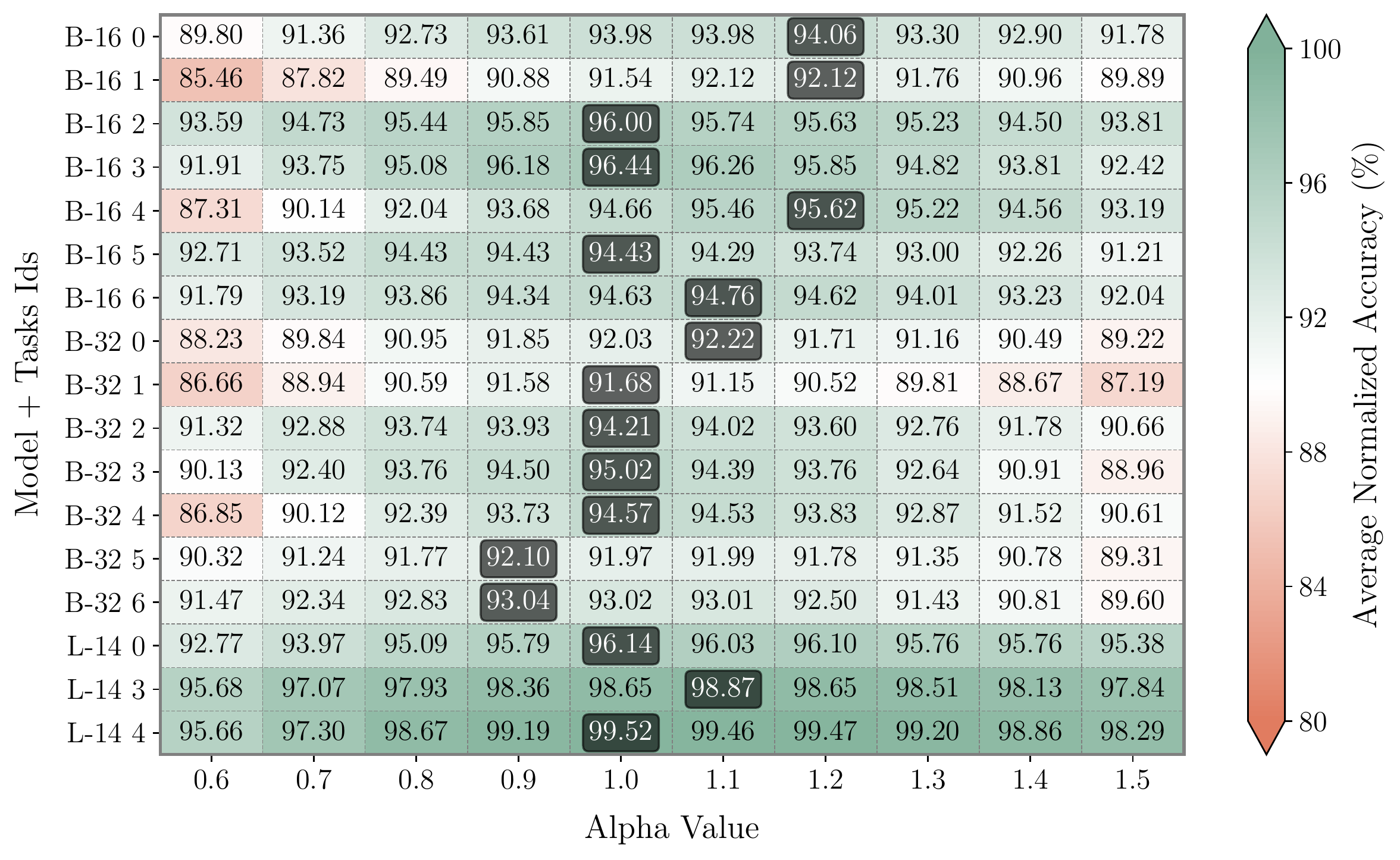}
  \caption{Best average normalized accuracy for different alpha values. The vertical labels indicate different sets of 8 tasks.}
  \label{fig:alpha}
\end{figure}

The aggregation in \cref{eq:merge_weights_layer} involves a scaling coefficient $\alpha$, typically tuned using a validation set, that can have a significant impact on the accuracy of the final model. However, as shown in \cref{fig:alpha}, our approach consistently achieves the best results with $\alpha = 1.0$, indicating that no additional scaling is necessary. Although purely empirical, this finding spares further tuning and validation data, thereby enhancing the practicality and efficiency of our method.

%%%%%%%%%%%%%%%%%%%%%%%%%%%%%%%%%%%%%%%%%%%%%
% TASK INTERFERENCE VS NORMALIZED ACCURACY
%%%%%%%%%%%%%%%%%%%%%%%%%%%%%%%%%%%%%%%%%%%%%

\subsection{Effect of task interference reduction}
\begin{figure}
  \centering
  \includegraphics[width=0.85\linewidth]{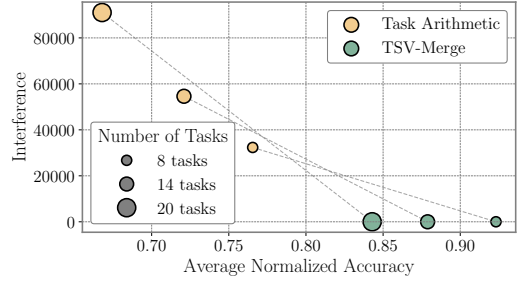}
  \caption{Singular Task Interference (STI) and average normalized accuracy for \baseline{Task Arithmetic} and \method{TSV-Merge} on the \model{ViT-B-32} model, evaluated across merges of 8, 14, and 20 tasks.}
  \label{fig:correlation_interference_accuracy}
\end{figure}

We report in \cref{fig:correlation_interference_accuracy} the interference of a \model{ViT-B-32} model both before and after the application of our \method{TSV-M} pipeline, assessed across three distinct sets of tasks with different cardinalities. In all instances, we observed that a reduction in interference is closely associated with a significant gain in accuracy. This provides empirical evidence for the effectiveness of our approach in minimizing interference and, ultimately, enhancing the merging process. 

\section{Conclusions}
%  BROAD CONTEXT AND HIGH-LEVEL RECAP
In this paper, we tackled the problem of model merging by analyzing the SVD of per-layer task matrices, confirming their inherent low-rank structure and leveraging their singular vectors to define task interference.

% COMPRESS
Building on the former insight, we introduced \method{TSV-Compress} (\method{TSV-C}), a model compression algorithm that reduces task vectors to 10\% of their original size while retaining 99\% of the original accuracy. Unlike existing methods, our approach maintains a constant storage requirement, independent of the number of tasks.
% MERGE
We then developed \method{TSV-Merge} (\method{TSV-M}), combining low-rank approximation with interference reduction to create a novel model merging technique. Achieving normalized accuracy up to 97\%, our method offers a storage-efficient alternative to MoEs, ensembles, and multi-task learning.

% EVALUATION
Our extensive evaluation revealed that while interference reduction contributes most to performance improvements, the optimal results are achieved in combination with compression. We also observed that task interference decreases in deeper layers, and importantly, that our method does not require tuning a scaling coefficient.

% FUTURE WORK
Exploring alternative methods for task importance and rank approximation could be a valuable direction for future work. Currently, we use a uniform rank across tasks, approximating each task vector to \(\frac{1}{\ntasks}\) of the layer’s dimensionality. However, techniques like those from \citet{idelbayev2020low} or optimal hard thresholds from \citet{6846297} could be applied to select the optimal rank for each task individually.

\FloatBarrier

\subsubsection*{Acknowledgments}
This work was partially supported by PRIN 2022 project 20227YET9B ``AdVVent'' CUP code B53D23012830006 and by projects FAIR (PE0000013) and SERICS (PE00000014) under the MUR National Recovery and Resilience Plan funded by the European Union - NextGenerationEU.  This work was moreover partially supported by the project NEREO (Neural Reasoning over Open Data) project funded by the Italian Ministry of Education and Research (PRIN) Grant no. 2022AEFHAZ.

\FloatBarrier

%\newpage 

{
    \small
    \bibliographystyle{ieeenat_fullname}
    \bibliography{main}
}

\FloatBarrier
\clearpage
\newpage

\appendix
\section{Illustrative example: merging two tasks with rank-1 approximation}
Consider merging two distinct tasks by selecting only the first singular vector and singular value from the SVD for each task. This setting yields the following setup for each layer $L_i$:
$$\left[
\begin{array}{@{}c c@{}}
  \mid & \mid \\
  u_{1_{L_i}} & u_{2_{L_i}} \\
  \mid & \mid \\
\end{array}
\right]
\left[
\begin{array}{c c}
   \sigma_{1_{L_i}} & 0 \\
%\hline
   0 & \sigma_{2_{L_i}}
\end{array}
\right]
\left[
\begin{array}{ccc}
 - & v^{T}_{1_{L_i}} & - \\
%\hline
 - & v^{T}_{2_{L_i}} & - \\
\end{array}
\right]$$
In this formulation, $u_{1_{L_i}}$ originates from task 1 and $u_{2_{L_i}}$ from task 2, with analogous assignments for the singular vectors $v$ and singular values $\sigma$.

To elucidate the interaction between tasks, we examine three distinct cases, considering a single layer, thereby omitting the layer index $L_i$:

\begin{enumerate}
%ORTHOGONAL SV
\item \textbf{Orthogonal Singular Vectors:} when $u_{1}$ and $u_{2}$ (respectively $v$) are orthogonal, the similarity matrix $U^\top U$ (respectively $V^\top V$) is given by: 
$$\left [
\begin{array}{c c}
   1 &  0 \\
%\hline
   0 & 1
\end{array}
\right]
$$
The zeroes in the off-diagonal elements indicate no interference between the tasks. Consequently, the orthogonal components derived from different tasks operate independently, ensuring that each task does not affect the other.

%COLLINEAR SV
\item \textbf{Collinear Singular Vectors:} when $u_{1}$ and $u_{2}$ (respectively $v$) are collinear, either aligned in the same direction (angle of 0 degrees) or in the opposite direction (angle of 180 degrees), the similarity matrix $U^\top U$ (respectively $V^\top V$) takes the form: 
$$\left [
\begin{array}{c c}
   1 & \langle u,\pm u \rangle\\
%\hline
  \langle \pm u, u \rangle & 1
\end{array}
\right] 
$$
If the singular vectors are perfectly aligned (0 degrees), then $u_{1} = u_{2} = u$, simplifying the diagonal elements to $\langle u, u \rangle = \lVert u \rVert^2 =1$. Conversely, if the singular vectors are oppositely aligned (180 degrees), $u_1 = -u_2$, resulting in $\langle u,-u \rangle = -1$ Thus, the similarity matrices becomes:
$$\left [
\begin{array}{c c}
   1 & \pm 1 \\
%\hline
  \pm 1 & 1
\end{array}
\right]
$$
This structure reveals complete interference between the tasks: a double scaling effect when the vectors agree and complete cancellation when they disagree.

%PARTIALLY COLLINEAR
\item \textbf{Partially Collinear Singular Vectors:} when $u_{1}$ and $u_{2}$ (respectively $v$) are partially collinear, with the angle between them ranging from slightly greater than 0 degrees to less than 90 degrees or slightly more than 90 degrees to less than 180 degrees, similarity matrices expressed as:
\begin{align*}
\left [
\begin{array}{c c}
   1 & \langle u_1, u_2 \rangle \\
%\hline
  \langle u_2, u_1 \rangle &
   1
\end{array}
\right]
\end{align*}
In this case, the overlap between singular vectors induces a partial interaction between the tasks. The degree of interference, whether it is constructive or destructive, is proportional to the cosine of the angle between the singular vectors. This partial collinearity leads to subtle interplay, where the tasks influence each other to a degree dictated by their vector alignment.
\end{enumerate}

This example underscores the critical role of singular vector alignment in model merging, highlighting how orthogonality ensures independent task performance, collinearity leads to maximal interference and partial collinearity results in an intermediate level of task interaction.

\begin{figure*}[btp]
  \centering
  \includegraphics[width=\linewidth]{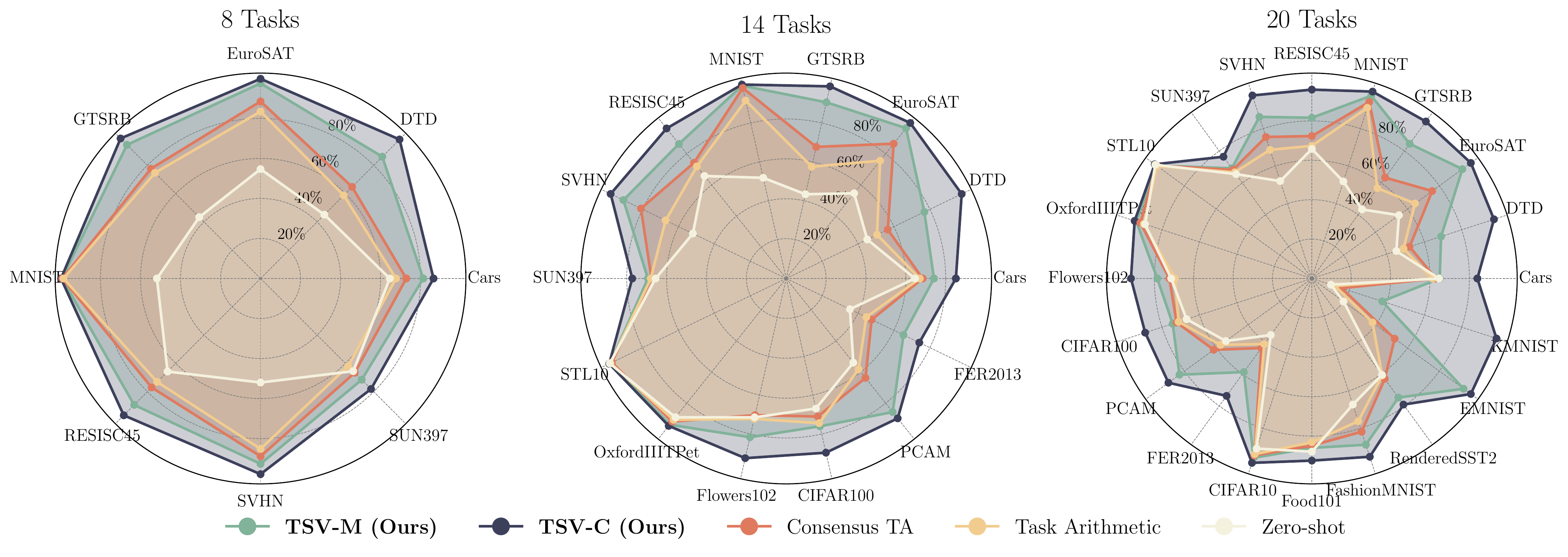}
  \caption{Absolute accuracy of a \model{ViT-B-16} merged over 8, 14, and 20 tasks, respectively.}
  \label{fig:radar_chart_B-16}
\end{figure*}

\section{Additional details}
\subsection{Implementation details and computational resources}

\paragraph{Normalized Accuracy}
\label{sec:normalized_accuracy}
To address the varying difficulties of the task, we report both normalized and absolute accuracies in our results. The normalized accuracy provides a relative performance metric by comparing the accuracy of the multi-task model to that of individually fine-tuned models. Specifically, the normalized accuracy is calculated as:
\begin{equation}
    \text{Normalized Accuracy} = \frac{1}{\ntasks} \sum_{i=1}^{\ntasks} \frac{\text{accuracy}(\theta_{MT}, t_i)}{\text{accuracy}(\theta_{ft_i}, t_i)}
\end{equation}
where $\ntasks$ is the total number of tasks, $\theta_{MT}$ represents the multi-task model and $\theta_{ft_i}$ denotes the individually fine-tuned model for task $t_i$. This metric allows for a more fair comparison by adjusting for the baseline performance of each task.

\paragraph{Datasets for tasks}
All benchmarks were performed by integrating the codebase provided by \citet{wang_localizing_2024}. In line with the principles of PEFT, we reused the already existing model checkpoints in the codebase for both the models and classification heads without additional fine-tuning. 
% This approach emphasizes the enhancement of performance by leveraging pre-existing resources, thus minimizing computational costs since the pre-trained and fine-tuned models are assumed to be available from an online repository (model zoos).

\begin{figure*}[!htbp]
  \centering
  \includegraphics[width=\linewidth]{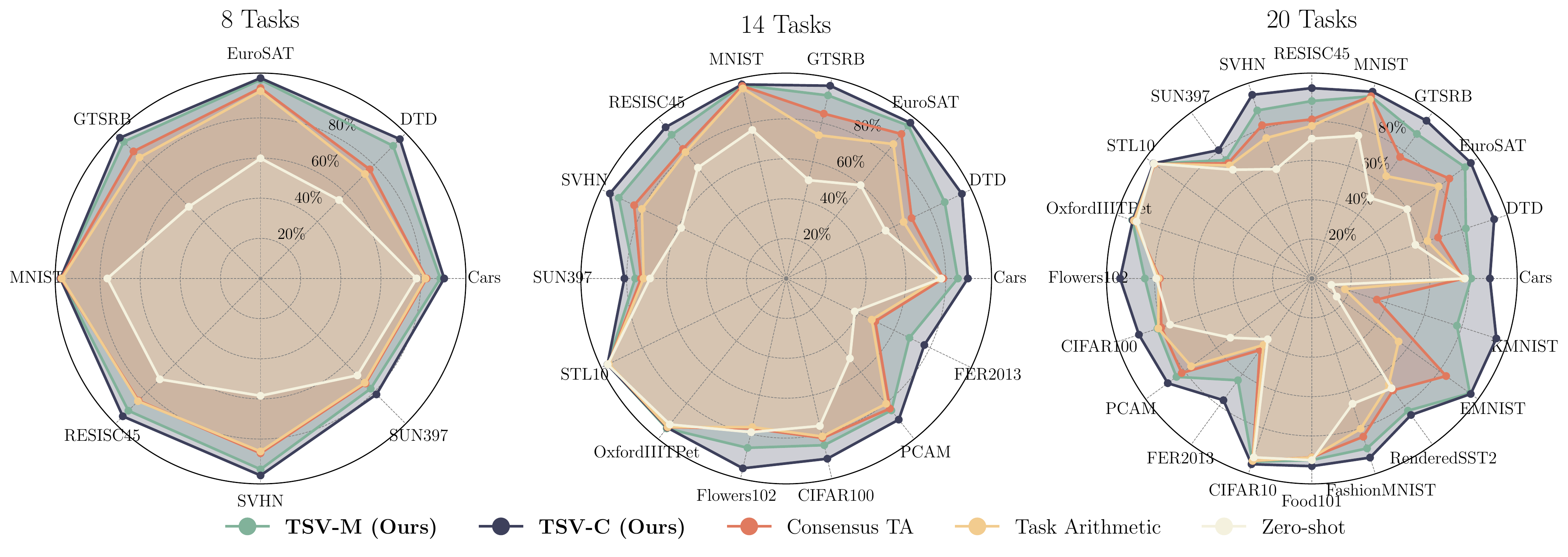}
  \caption{Absolute accuracy of a \model{ViT-L-14} merged over 8, 14, and 20 tasks, respectively.}
  \label{fig:radar_chart_L-14}
\end{figure*}

\paragraph{Implementation}
Our method utilizes the SVD, a matrix decomposition technique applicable to two-dimensional matrices.
For layers that are not represented as matrices (e.g., normalization layers) we default to standard \method{Task Arithmetic}. In particular, we employ Knut's algorithm \citep{doi:10.1080/00401706.1962.10490022} to compute the average efficiently. This ensures that all fine-tuned model task layers, regardless of their structure, are appropriately integrated into the merged model.

\paragraph{Compute Resources}
We utilize PyTorch as deep learning framework. All the merging and evaluations were conducted on an \texttt{NVIDIA 4060Ti} GPU with 16GB of memory, and an \texttt{Intel i7-6800K} CPU equipped with 64GB of RAM. For experiments that need more than 64GB of RAM, we resort to a shared \texttt{HTCondor} cluster equipped with \texttt{NVIDIA P6000} GPUs.

\subsection{Hyperparameter Settings}
Following \texttt{Task Arithmetic} \cite{ilharco_editing_2023} and \texttt{Consensus TA} \cite{wang_localizing_2024}, we apply a single scaling factor, $\alpha$, to adjust the multi-task vector within the model merging techniques outlined in \Cref{tab:task_acc}. This scaling factor is optimized, when feasible, over the range \{0.0, 0.1, ..., 2.9, 3.0\}, with the optimal value selected based on the average validation performance across all tasks. 
However, as discussed in \Cref{sec:alpha}, our experimental findings indicate that the proposed \method{TSV-Merge} method does not strictly depend on this hyperparameter, as the marginal performance gains from tuning $\alpha$ do not justify the computational resources required for the evaluation on the validation datasets. Consequently, this allows us to eliminate the necessity for validation datasets and the corresponding labels from the prerequisites of the method, further simplifying the practicality and resource usage of the approach. The evaluation is performed on batch of 32 images. To produce \cref{fig:alpha} we selected the following subsets of 8 tasks from the 20 available, using the whitening method to speed up computation, the number in the image is the index indicating the subset: 
\begin{enumerate}
\setcounter{enumi}{-1}

    \item  \dataset{SVHN}, \dataset{SUN397}, \dataset{STL10}, \dataset{OxfordIIITPet}, \dataset{Flowers102}, \dataset{CIFAR100}, \dataset{PCAM}, \dataset{FER2013}

    \item  \dataset{PCAM}, \dataset{FER2013}, \dataset{CIFAR10}, \dataset{Food101}, \dataset{FashionMNIST}, \dataset{RenderedSST2}, \dataset{EMNIST}, \dataset{KMNIST}

    \item  \dataset{EuroSAT}, \dataset{GTSRB}, \dataset{MNIST}, \dataset{RESISC45}, \dataset{SVHN}, \dataset{SUN397}, \dataset{STL10}, \dataset{OxfordIIITPet}

    \item \dataset{Cars}, \dataset{DTD}, \dataset{EuroSAT}, \dataset{GTSRB}, \dataset{MNIST}, \dataset{RESISC45}, \dataset{SVHN}, \dataset{SUN397}

    \item  \dataset{Cars}, \dataset{DTD}, \dataset{EuroSAT}, \dataset{GTSRB}, \dataset{FashionMNIST}, \dataset{RenderedSST2}, \dataset{EMNIST}, \dataset{KMNIST}

    \item  \dataset{MNIST}, \dataset{RESISC45}, \dataset{SVHN}, \dataset{SUN397}, \dataset{STL10}, \dataset{OxfordIIITPet}, \dataset{Flowers102}, \dataset{CIFAR100}

    \item  \dataset{STL10}, \dataset{OxfordIIITPet}, \dataset{Flowers102}, \dataset{CIFAR100}, \dataset{PCAM}, \dataset{FER2013}, \dataset{CIFAR10}, \dataset{Food101}

\end{enumerate} 

\begin{figure}[tbp]
  \centering
  \includegraphics[width=\linewidth]{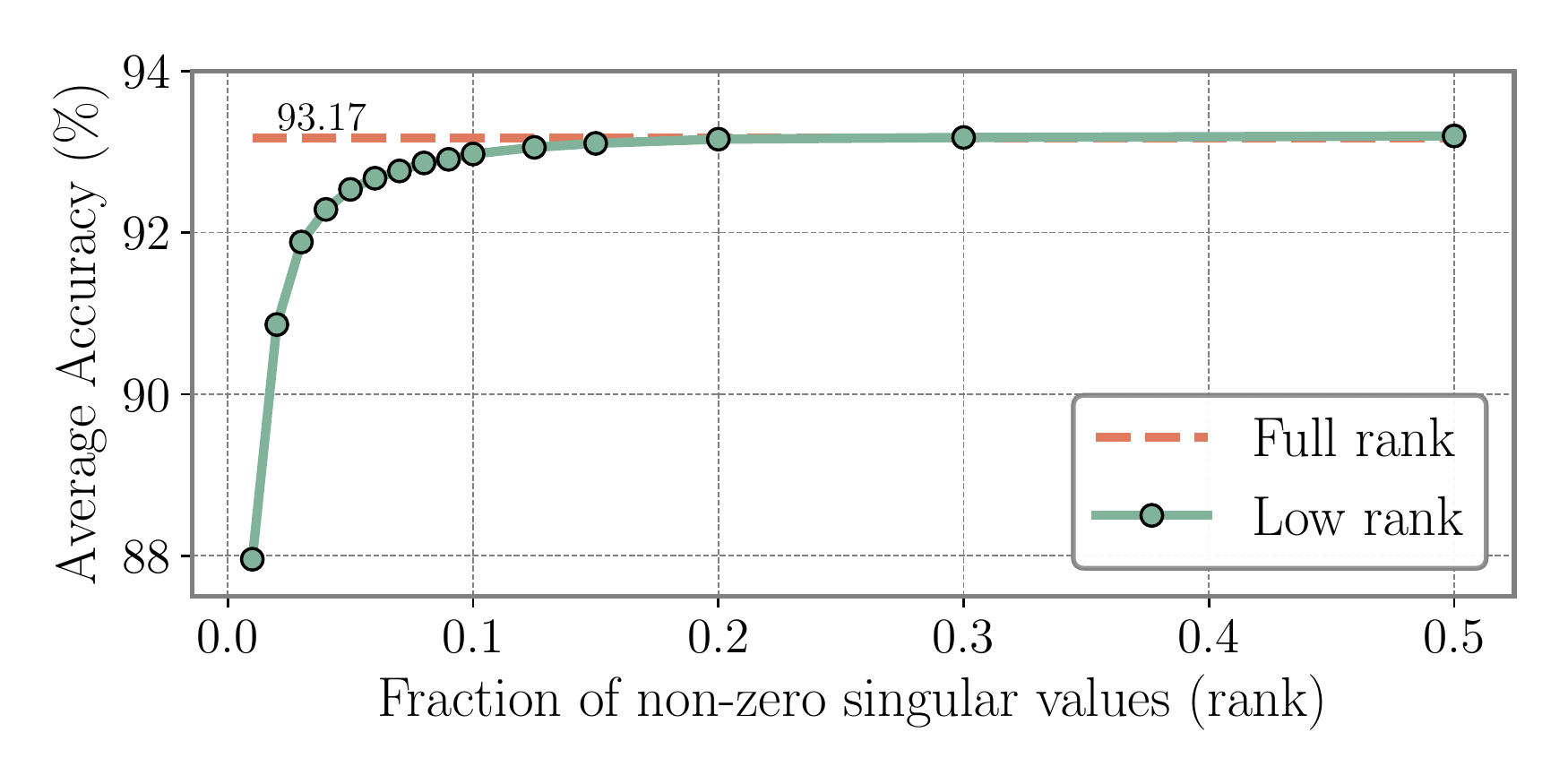}
    \caption{Mean absolute accuracy of the \model{ViT-B-16} model across increasing fractions of retained singular components, averaged over 20 tasks. The red line represents the average accuracy of the original fine-tuned models with full-rank task matrices, while the green line shows the accuracies using low-rank approximations.}
  \label{fig:acc_by_rank_B-16}
\end{figure}

\subsection{Storage cost calculation}

\begin{figure}[tbp]
  \centering
  \includegraphics[width=\linewidth]{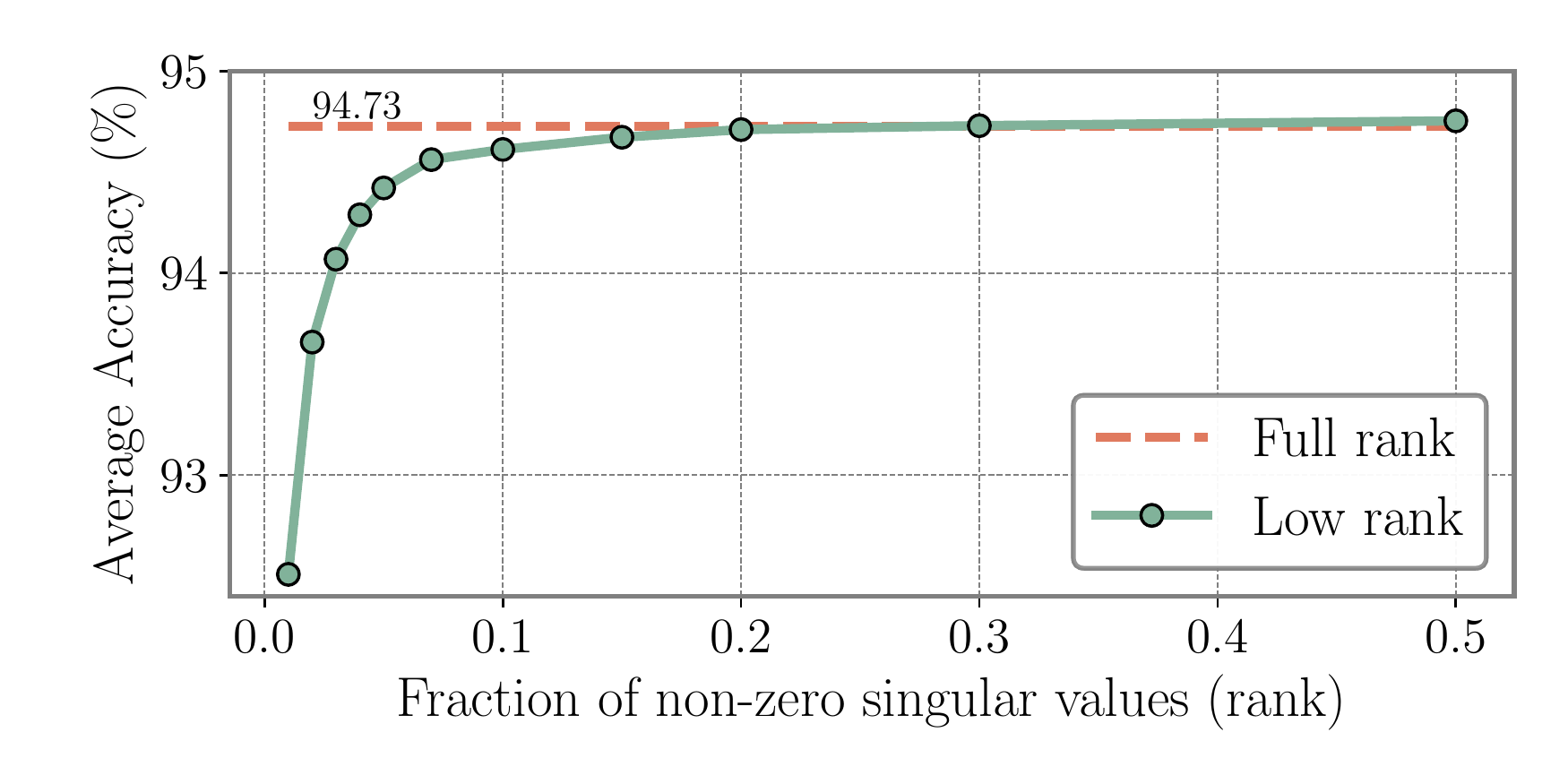}
    \caption{Mean absolute accuracy of the \model{ViT-L-14} model across increasing fractions of retained singular components, averaged over 20 tasks. The red line represents the average accuracy of the original fine-tuned models with full-rank task matrices, while the green line shows the accuracies using low-rank approximations.}
  \label{fig:acc_by_rank_L-14}
\end{figure}

Suppose we have a neural network comprising of $L$ two-dimensional layers, each of dimension $d \times m$, and $N$ one-dimensional layers of size $c$. The total number of parameters in the network is therefore: 
$$\text{Params(NN)} = L \times (d \times m) + N \times c.$$ 
In standard \method{Task Arithmetic}, one must store the same number of parameters to obtain a task vector. In contrast, our approach provides the flexibility to select the number of parameters to preserve based on storage constraints or the desired needed performance, to adhere to the chosen constraints.
Under the above assumptions, our method applies the truncated SVD to each two-dimensional layer. This decomposition yields two matrices of singular vectors, $U$ and $V$, and a vector of singular values, $\sigma$, specifically:
\begin{itemize}
    \item $U$ of size $d \times k$,
    \item $V$ of size $k \times m$,
    \item $\sigma$ of size $k$,
\end{itemize} 
where $k = \text{min}(d,m)$. We select a reduced rank $k' \ll k$ to approximate each layer's task matrix. Consequently, the total number of parameters for TSV becomes:
$$\text{Params(TSV)} = L \times ((d \times k') + k' + (k' \times m)) + N \times c$$
To demonstrate that our method results in fewer stored parameters than the original parameter count, we require that $k'<\frac{d \times m}{d+m+1}$. This condition ensures: $$\text{Params(NN)} > \text{Params(TSV)}$$ Substituting the expressions, yields: $$L \times (d \times m) + N \times c > L \times ((d \times k') + k' + (k' \times m)) + N \times c$$
Simplifying, we obtain:
\begin{equation}
    \begin{split}
        L \times (d \times m) &> L \times ((d \times k') + k' + (k' \times m)) \\ 
        (d \times m) &> ((d \times k') + k' + (k' \times m)) \\
        d \times m &> k' \times (d  + 1 + m) \\
        k' &< \frac{d \times m}{d+m+1}. 
    \end{split}
\end{equation}
This inequality confirms that our method reduces the storage requirements of a task vector when $k'<\frac{d \times m}{d+m+1}$. Empirical evidence from \Cref{fig:acc_by_rank,fig:acc_by_rank_B-16,fig:acc_by_rank_L-14} suggests that selecting $k'< 0.1 \times k = 0.1\times\text{min}(d,m)$ is sufficient to preserve most of the task performance, preserving the main requirement of performance. Furthermore, it is easy to prove that when choosing $k'=\frac{k}{\ntasks}$, the inequality is always satisfied for $\ntasks \ge 3$, respecting the main requirement of limited storage usage.
\begin{table}[b]
  \centering
  \resizebox{1\columnwidth}{!}{
  \begin{tabular}{ccccc}
    \toprule
    \vspace{-0.1cm}Low-rank       & Interf.  & \multicolumn{3}{c}{\model{ViT-B-16}}                                           \\
    \cmidrule{3-5}
    approx. & reduction           & 8 tasks                                  & 14 tasks                & 20 tasks                \\
    \midrule
    \negxmark     & \negxmark     & 79.6 (+0.0)                         & 75.9 (+0.0)        & 70.8 (+0.0)       \\
    \poscheckmark & \negxmark     & 79.6 (+0.0)                         & 74.9 \worse{1.0}   & 70.0 \worse{0.8}   \\
    \negxmark     & \poscheckmark & 84.8 \improv{5.2}                   & 79.0 \improv{4.1}  & 73.2 \improv{3.2}  \\
    \poscheckmark & \poscheckmark & 93.9 \improv{9.1}                   & 91.0 \improv{12.0} & 86.5 \improv{13.3} \\
    \bottomrule
  \end{tabular}
  }
  \caption{Comparison of different versions of \baseline{Task Arithmetic}, comprising either the low-rank approximation step, the interference reduction step, or both. The method performing both corresponds to the proposed \method{TSV-Merge}.}
  \label{tab:ablation-study-B-16}
\end{table}
\section{Proofs}
We hereby prove the claims outlined in the main manuscript.

%\subsection{\texorpdfstring{Extension of \Cref{prop:whitening-procrustes}}{Extension of Proposition 4.1}}
%Here we show a step-by-step version of the derivation for \Cref{prop:whitening-procrustes}:
%\begin{proposition}
%    The whitening transformation $X \mapsto  X(X^\top X)^{-\frac{1}{2}}$ and the orthogonal Procrustes transformation $X \mapsto PQ^\top$, where $X = PDQ^\top$ is the SVD of $X$, are equivalent.
%\end{proposition}
%\begin{proof}

%     Given the SVD: $X=PDQ^\top$.
    
%     Compute $(X^\top X)^{-\frac{1}{2}}$:
%     \begin{equation}
%     \begin{split}        
%         (X^\top X)^{-\frac{1}{2}} &= (QD^{\top}P^{\top}PDQ^{\top})^{-\frac{1}{2}} \\
%         &= (QD^{\top}IDQ^{\top})^{-\frac{1}{2}} \\
%         &=(QD^{2}Q^{\top})^{-\frac{1}{2}} \\ 
%         &= Q D^{-1} Q^\top.
%     \end{split}
%     \end{equation}
%     Recall that since $D$ is diagonal $D^\top D=D^2$ and $P$ and $Q$ are orthogonal matrices ($P^{\top}P=I$).
    
%     Therefore the whitening transformation can be rewritten as: 
%     \begin{equation}
%     \begin{split}  
%         X (X^\top X)^{-1/2} &= (PDQ^\top)(Q D^{-1} Q^\top) \\
%         &=PDID^{-1}Q^\top \\
%         &=PQ^\top
%     \end{split}
%     \end{equation}
%     completing the proof.
%     Thus, the whitening transformation yields the same result as the orthogonal Procrustes transformation.
% \end{proof}

\subsection{Characterization of the similarity matrices}

\begin{proposition} \label{prop:psd}
    The matrix \( \hat{U}^\top \hat{U} \) is positive definite.
\end{proposition}

\begin{proof}
    We define \( \hat{U}^\top \hat{U} \), where \( \hat{U} \) is a generic \( d \times k \) rectangular matrix. Consequently, \( \hat{U}^\top \hat{U} \) is a \( k \times k \) square matrix. To establish that \( \hat{U}^\top \hat{U} \) is positive definite, it suffices to demonstrate that for all non-zero vectors \( x \in \mathbb{R}^{k} \), the following inequality holds:
    \[
    x^\top \hat{U}^\top \hat{U} x > 0.
    \]
    
    This expression can be rewritten as:
    \[
    x^\top (\hat{U}^\top \hat{U}) x = (\hat{U} x)^\top (\hat{U} x) = \lVert \hat{U} x \rVert^2.
    \]
    Here, \( \lVert \hat{U} x \rVert^2 \) denotes the squared Euclidean norm of the vector \( \hat{U} x \), which is always non-negative. Moreover, assuming that \( \hat{U} \) has full column rank, the norm \( \lVert \hat{U} x \rVert^2 \) is strictly positive for any non-zero vector \( x \). Therefore, we have:
    \[
    \lVert \hat{U} x \rVert^2 > 0 \quad \text{for all} \quad x \in \mathbb{R}^{k}, \, x \neq 0.
    \]
    
    This implies that:
    \[
    x^\top \hat{U}^\top \hat{U} x > 0 \quad \text{for all} \quad x \in \mathbb{R}^{k}, \, x \neq 0,
    \]
    which confirms that \( \hat{U}^\top \hat{U} \) is positive definite.
\end{proof}

\begin{corollary} \label{cor:psd_invertible}
    Since \( \hat{U}^\top \hat{U} \) is positive definite, then \( \hat{U}^\top \hat{U} \) is invertible.
\end{corollary}

\begin{proof}
    From \Cref{prop:psd}, we have established that \( \hat{U}^\top \hat{U} \) is a positive definite matrix. A positive definite matrix, by definition, has all its eigenvalues strictly positive. Let \( \lambda_1, \lambda_2, \ldots, \lambda_k \) denote the eigenvalues of \( \hat{U}^\top \hat{U} \). Therefore, we have:
    \[
    \lambda_i > 0 \quad \text{for all} \quad i = 1, 2, \ldots, k.
    \]
    
    The determinant of \( \hat{U}^\top \hat{U} \) is the product of its eigenvalues:
    \[
    \det(\hat{U}^\top \hat{U}) = \prod_{i=1}^{k} \lambda_i.
    \]
    Since each \( \lambda_i \) is positive, their product is also positive:
    \[
    \det(\hat{U}^\top \hat{U}) > 0.
    \]
    
    A matrix is invertible if and only if its determinant is non-zero. Given that \( \det(\hat{U}^\top \hat{U}) > 0 \), it follows that \( \hat{U}^\top \hat{U} \) is invertible.
    
    Therefore, \( \hat{U}^\top \hat{U} \) is invertible.
\end{proof}

\subsection{Observations}

Since \( \hat{U}^\top \hat{U} \) is a real symmetric matrix, it admits an eigendecomposition of the form
\[
\hat{U}^\top \hat{U} = Q \Lambda Q^{-1} = Q \Lambda Q^\top,
\]
where:
\begin{itemize}
    \item \( \Lambda \) is a diagonal matrix containing the real eigenvalues of \( \hat{U}^\top \hat{U} \),
    \item \( Q \) is an orthogonal matrix whose columns are the orthonormal eigenvectors of \( \hat{U}^\top \hat{U} \), satisfying \( Q^\top = Q^{-1} \).
\end{itemize}

The inverse of \( \hat{U}^\top \hat{U} \) exists (see \Cref{cor:psd_invertible}), and can be expressed using its eigendecomposition as
\[
(\hat{U}^\top \hat{U})^{-1} = Q \Lambda^{-1} Q^{-1} = Q \Lambda^{-1} Q^{\top}.
\] 
Additionally, since \( \Lambda \) is a diagonal matrix with non-zero diagonal entries (\Cref{prop:psd}), its inverse \( \Lambda^{-1} \) is straightforward to compute, with each diagonal element given by
\[
\Lambda^{-1} = \text{diag}\left( \frac{1}{\lambda_i} \right),
\]
where \( \lambda_i \) are the eigenvalues of \( \hat{U}^\top \hat{U} \).

Furthermore, the eigenvalues of \( (\hat{U}^\top \hat{U})^{-1} \) are \( \frac{1}{\lambda_i} \), each of which is positive since \( \lambda_i > 0 \) for all \( i \) (following by the definition in \Cref{prop:psd}). Consequently, \( (\hat{U}^\top \hat{U})^{-1} \) is also a positive definite matrix.

These observations confirm that not only is \( \hat{U}^\top \hat{U} \) positive definite, but its inverse inherits this property due to the positivity of its eigenvalues.
\begin{table}[h!]
  \centering
  \resizebox{1\columnwidth}{!}{
  \begin{tabular}{ccccc}
    \toprule
    \vspace{-0.1cm}Low-rank & Interf.       & \multicolumn{3}{c}{\model{ViT-L-14}}                         \\
    \cmidrule{3-5}
    approx.                 & reduction     & 8 tasks            & 14 tasks           & 20 tasks           \\
    \midrule
    \negxmark               & \negxmark     & 88.6 (+0.0)        & 84.0 (+0.0)        & 78.1 (+0.0)        \\
    \poscheckmark           & \negxmark     & 87.9 \worse{0.7}   & 83.4 \worse{0.6}   & 77.2 \worse{0.9}   \\
    \negxmark               & \poscheckmark & 92.1 \improv{4.2}  & 86.8 \improv{3.4}  & 81.0 \improv{3.8}  \\
    \poscheckmark           & \poscheckmark & 97.0 \improv{4.9}  & 94.4 \improv{7.6}  & 92.5 \improv{11.5} \\
    \bottomrule
  \end{tabular}
  }
  \caption{Comparison of different versions of \baseline{Task Arithmetic}, comprising either the low-rank approximation step, the interference reduction step, or both. The method performing both corresponds to the proposed \method{TSV-Merge}.}
  \label{tab:ablation-study-L-14}
\end{table}
\subsection{\texorpdfstring{Proof of \Cref{theo:error}}{Proof of Theorem 6.1}}
\label{mini_proof}
\setcounter{theorem}{0}
\renewcommand{\thesection}{6}
\begin{theorem}
Let \(T \in \mathbb{N}\) such that \(T > 4\). Define \( U = [U_1, \dots, U_T] \) as the matrix obtained by concatenating \(T\) orthogonal matrices \( U_i \), each of shape \(n \times n\). Let \( \widehat{U} = [\widehat{U}_1, \dots, \widehat{U}_T] \) be the matrix formed by truncating each \( U_i \) to its first \(k\) columns. Denote by \(X\) and \(\widehat{X}\) the matrices resulting from Procrustes orthonormalization of \(U\) and \(\widehat{U}\), respectively. If \(k \leq n \frac{T - 2\sqrt{T}}{T}\), then 
    \[
    \| U - X \|_F \geq \| \widehat{U} - \widehat{X} \|_F \,.
    \]
\end{theorem}

\begin{proof}
 Let us consider the SVD decomposition of \(U\) and \(\widehat{U}\): \(U = P_u \Sigma_u P_v \) and \(\widehat{U} = R_u \widehat{\Sigma_u} R_v\). 
  \(X\) and \( \widehat{X} \)  obtain as \( X = P_u P_v^\top , \; \widehat{X} = R_u R_v^\top \) respectively 
We first consider the Frobenius norm of \(|| X- U||_F\). 
Notice that the singular values of \(U\) are the square root of the  eigenvalues of \( \Sigma_u = U U^\top \).

\(U U^\top = \sum_{i=1}^N U_i U_i^{\top} = T I_n.  \) As a consequence, the eigenvalues of \(U U^{\top} \) are all equal to \(T\) and the singular values are all equal to $\sqrt{T}$.
\begin{align*}
|| X- U||_F &= ||P_u P_v^\top  - P_u \Sigma_u P_v^\top||_F & \\
&= || P_u (I - \Sigma_u) P_v^\top ||_F  & \\
&= ||I_n - \Sigma_u ||_F & 
%(we \; used \; that \; P_u P_v^\top \; are \; orthogonal )
\\
&= ||I_n- \sqrt{T} I_n||_F & 
%(we \; used \; that \; \Sigma_u = \sqrt{N} I_n)
\\
&= \sqrt{n} (\sqrt{T}-1).
\end{align*}

We are now left to compute \( ||\widehat{X}-\widehat{U}||_F \). In this case, we are not able to compute the exact norm without other assumptions, but we can provide an upper bound that gives us a sufficient condition to prove our statement. 
As before 
\begin{align*}
    ||\widehat{X}-\widehat{U}||_F &= ||I_n - \widehat{\Sigma}_u ||_F\\
    &= \sqrt{\sum_{i=1}^n (\hat{\sigma}_i -1)^2 }.
    \end{align*}
    where $\hat{\sigma}$ are the singular values of \( \widehat{U} \).

Notice that 
\( \sum_{i=1}^n \hat{\sigma}^2_i = \text{tr}(\widehat{U} \widehat{U}^{\top}) =  \text{tr}(\widehat{U}^{\top} \widehat{U} ) \) 
and \( \widehat{U}^{\top} U  \) is a \( Tk \times Tk \) matrices with diagonal elements equal to one, so \( \text{tr}(\widehat{U}^{\top} \widehat{U} ) = kT \).

%Moreover \( \widehat{U} \widehat{U}^{\top} \) is positive semi-definite, so the smallest singular value is larger or equal to 0, i.e. I can lower-bound the sum \( \sum_{i=1}^n \sigma_i \geq 0\). 

Moreover,  

\begin{align}
  \sum_{i=1}^n \hat{\sigma_i} &=  \sum_{i=1}^n  \sqrt{\lambda_i (\widehat{U} \widehat{U}^{\top})} \\
  & \geq   \sqrt{ \sum_{i=1}^n \lambda_i (\widehat{U} \widehat{U}^{\top})} \\
  &= \sqrt{\text{tr}(\widehat{U} \widehat{U}^{\top})} = \sqrt{T k}  
\end{align}

Notice that this upper bound is tight, indeed the sum of the singular values of $\hat{U}$ must lie within: \( \sqrt{kT} \leq \sum_{i=1}^n \hat{\sigma}_i \leq kT \). 
The minimum \( \sqrt{kT} \) is achieved if all matrices $U_i$ are equals, on the other end, the maximum \( kT \) is achieved if the \(kT \) columns are orthonormal. 

Putting everything together, 
\begin{align}
    || \widehat{X} - \widehat{U} ||_F  &= \sqrt{\sum_{i=1}^n (\hat{\sigma}_i -1)^2} & \\
    &= \sqrt{n + \sum_{i=1}^n \hat{\sigma}^2_i  - 2\sum_{i=1}^n \hat{\sigma}_i } & \\
    & = \sqrt{ n + kT  - 2\sum_{i=1}^n \hat{\sigma}_i } & \\
    & \leq \sqrt{ n + kT -2\sqrt{kT}     }.
\end{align}

So we have to check for what values of $k $ it holds that \(\sqrt{ n + kT - 2 \sqrt{kT}     } \leq \sqrt{n }(\sqrt{T}-1) \).

We have that 
\begin{equation}
    \sqrt{ n + kT - 2 \sqrt{kT}     } \leq  \sqrt{ n + kT    } \leq  \sqrt{n }(\sqrt{T}-1) 
    \label{eq:that_need_to_be_satistfied}
\end{equation}

Equation \ref{eq:that_need_to_be_satistfied} is satisifed if 

%Solving the equation for $k $ gives us that the inequality above is satisfied if \( k \leq \frac{\left( 1 + \sqrt{1 + nT - 2 n \sqrt{T}}\right)}{T} \). 

\( k \leq n \frac{T- 2 \sqrt{T}}{T}\).
This concludes the proof.
 Since $k $ is a positive number the inequaility of meaningful for  $T>4$.
\end{proof}

% \begin{remark}
% For example, when \(T = 10\) as in our case, the quantity \(\frac{T - 2\sqrt{T}}{T} = \frac{10 - 2\sqrt{10}}{10} \geq \frac{1}{3}\). 

% In our case, we choose \(k \approx \frac{n}{T}\). For \(T = 10\), this gives \(k \approx \frac{n}{10}\), which is indeed smaller than \(\frac{n}{3}\). Therefore, the condition \(k \leq \frac{1}{3}n\) holds, and we are perfectly fine.
% \end{remark}
%
\begin{figure}[h]
  \centering
  \includegraphics[width=\linewidth]{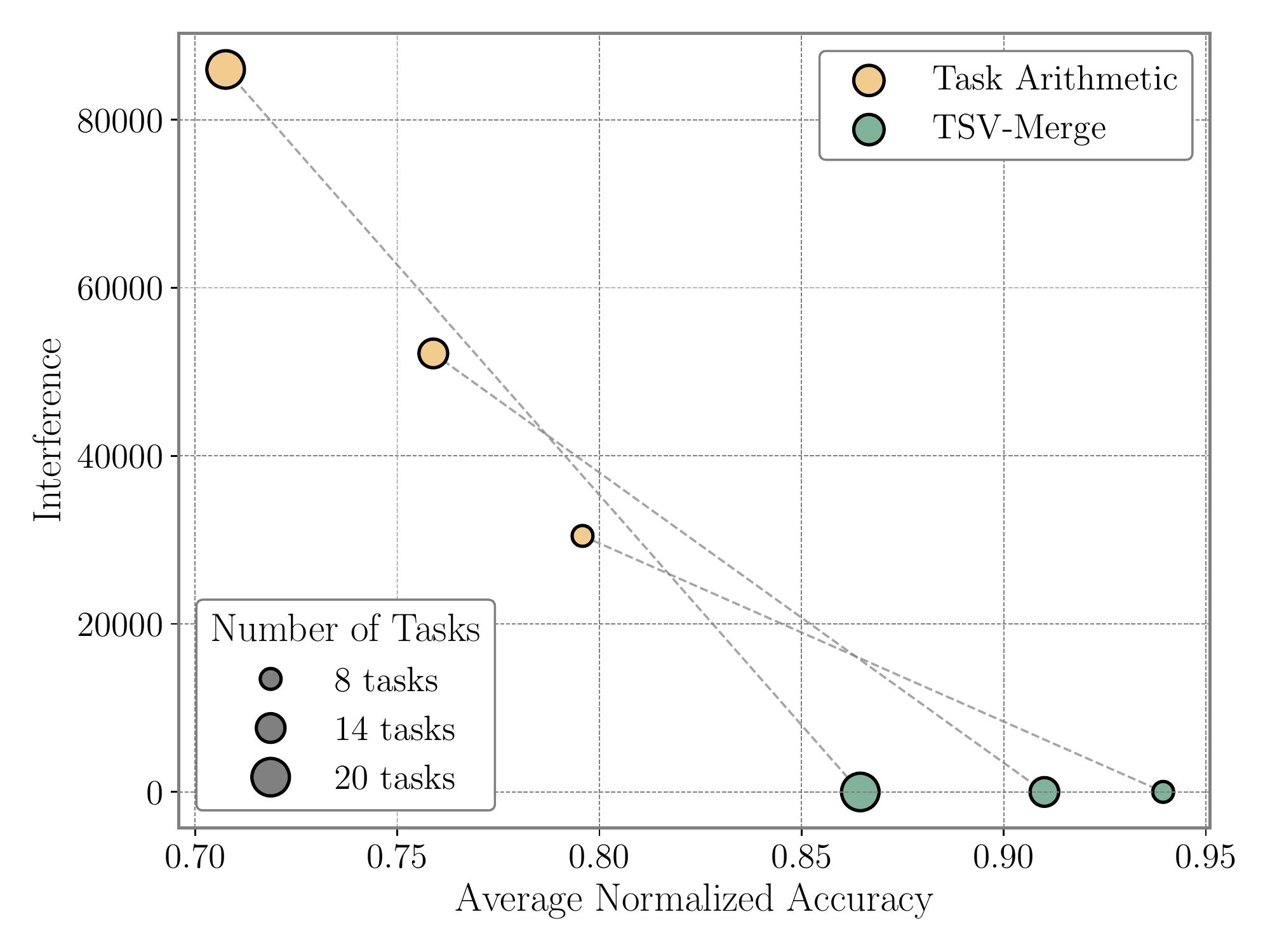}
  \caption{Singular Task Interference (STI) and average normalized accuracy for \baseline{Task Arithmetic} and \method{TSV-Merge} on the \model{ViT-B-16} model, evaluated across merges of 8, 14, and 20 tasks.}
  \label{fig:correlation_interference_accuracy_B-16}
\end{figure}
\begin{figure*}[!ht]
  \centering
  \includegraphics[width=\linewidth]{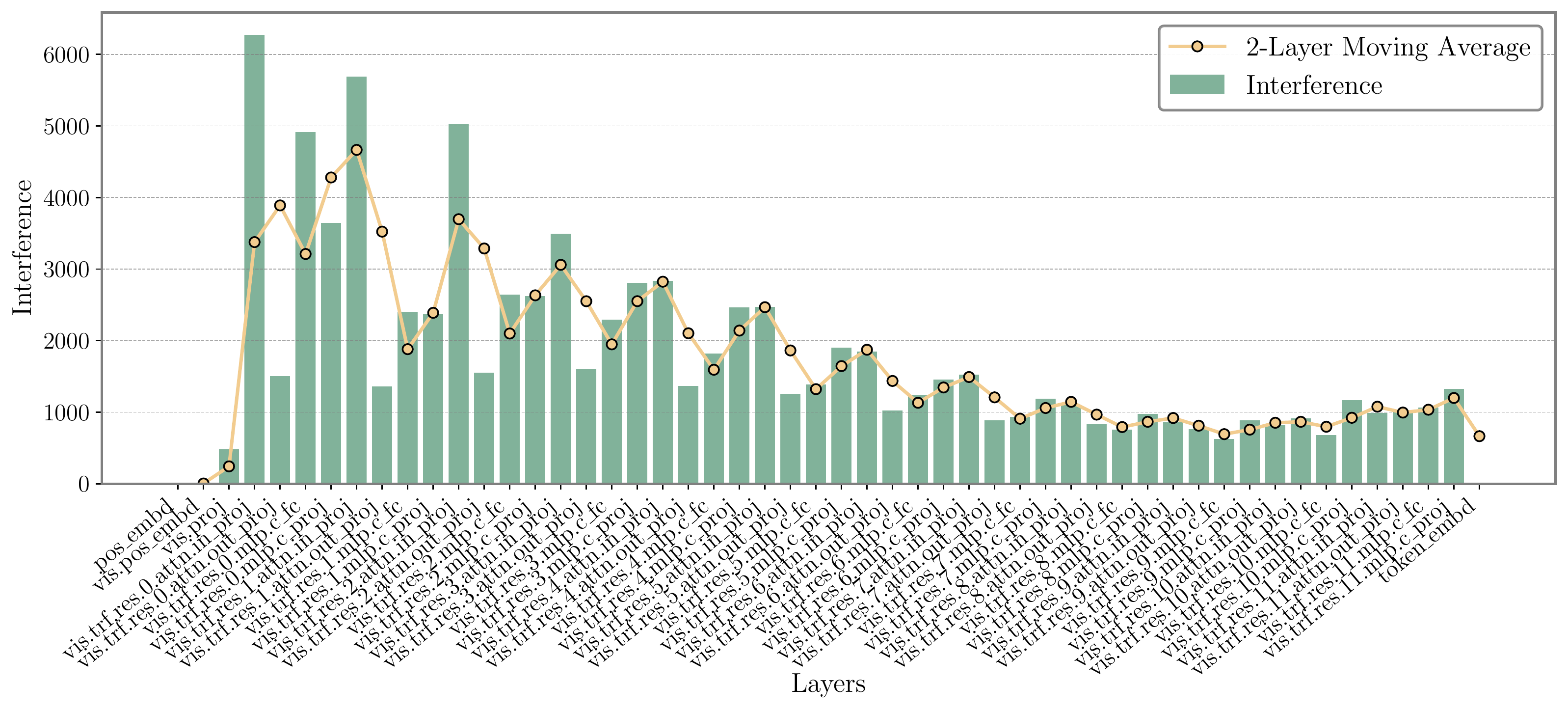}
  \caption{Detailed view of Singular Task Interference (STI) across layers in a \model{ViT-B-32} for 20 tasks. The interference trend is high in early layers and decreases later. Here, the pattern for each transformer block is observable, the interference first increases and then drops in each attention-out layer.}
  \label{fig:detailed_interference_by_layer}
\end{figure*}
\renewcommand{\thesection}{D}
\section{Additional experimental results}

\subsection{Per-dataset performance metrics}
In \Cref{sec:MM_results}, we present comprehensive results for individual tasks using the \model{ViT-B-32} model. Here we include analogous radar plots for the \model{ViT-B-16} model in \Cref{fig:radar_chart_B-16} and the \model{ViT-L-14} model in \Cref{fig:radar_chart_L-14}. The analyses of these models reveal findings consistent with those reported for \model{ViT-B-32} in the main text.

\subsection{Extended analysis}

\subsubsection{Whitening vs. SVD}
As we have seen in \Cref{subsec:tsv-merge}, applying a whitening transformation to the matrices of task singular vectors is mathematically equivalent to solving the Orthogonal Procrustes problem. However, implementing these two approaches may yield different results depending on the distinct matrix decomposition algorithms employed. In this study, we used PyTorch to compute both eigendecomposition and SVD, observing slightly different results that may be attributed to numerical errors. To more robustly compute the matrix square root for the eigendecomposition case, we compute 
$$\Lambda^{-\frac{1}{2}} = \text{diag}\left(\frac{1}{\sqrt{|\lambda_i|}+\epsilon}\right)$$ 
where $\epsilon = 1\mathrm{e}{-12}$ prevents division by $0$ and the absolute value avoids numerical errors producing small negative values in magnitude less than $1\mathrm{e}{-6}$.

\subsubsection{Impact of rank}
\begin{figure*}
  \centering
  \includegraphics[width=0.90\linewidth, height=0.94\textheight]{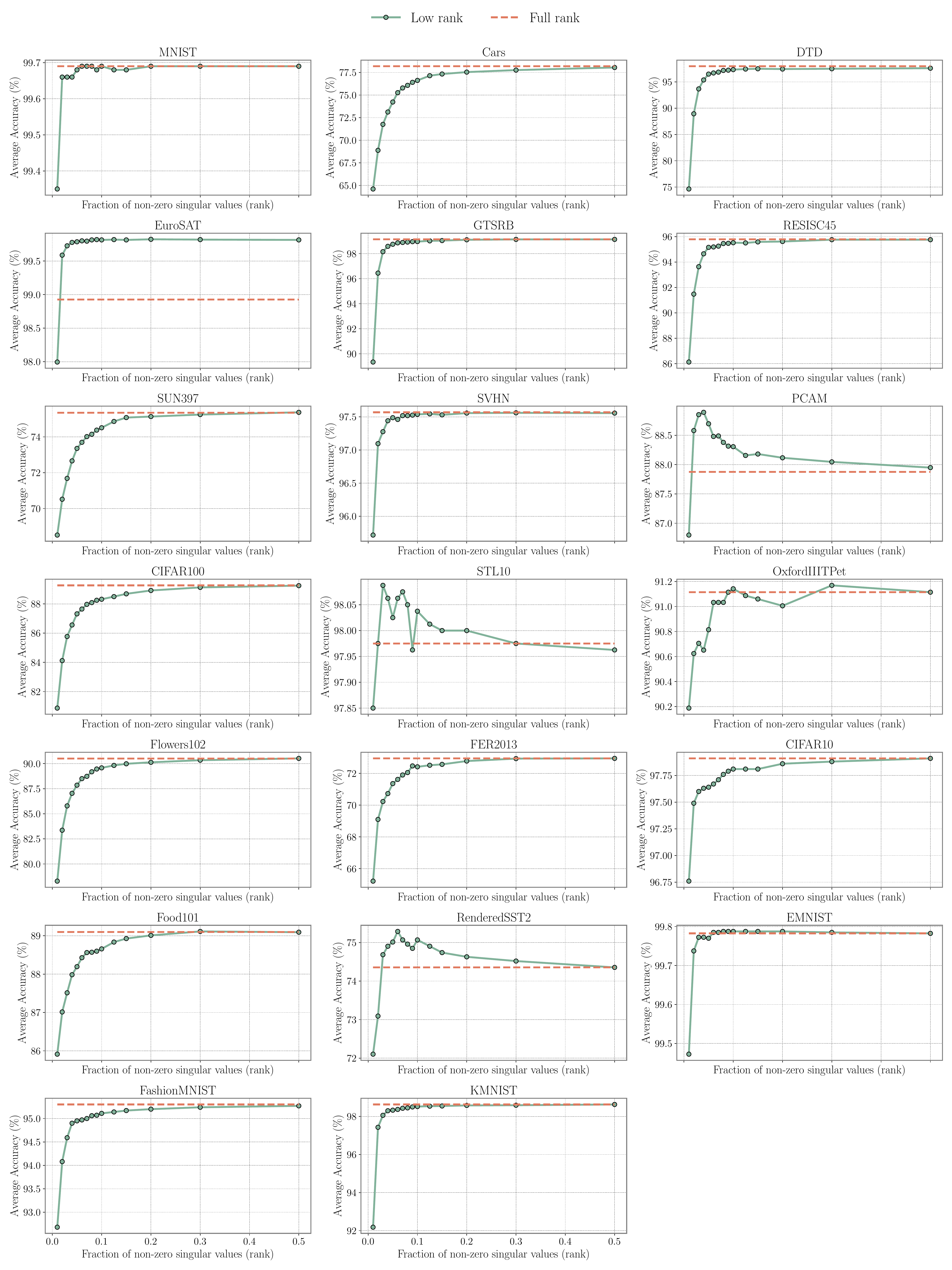}
  \caption{Absolute accuracy of the \model{ViT-B-32} model across increasing fractions of retained singular components, for each task. The red line represents the accuracy of the original fine-tuned models with full-rank task matrices, while the green line shows the accuracies using low-rank approximations.}
  \label{fig:rank_analysis_by_dataset}
\end{figure*}

The \Cref{subsec:low-rank} shows that the task matrices of a \model{ViT-B-32} are inherently low-rank and a small percentage of TSVs is enough to approximate each layer with satisfying results. We here provide the same plots for the models \model{ViT-B-16} (\Cref{fig:acc_by_rank_B-16}) and \model{ViT-L-14} (\Cref{fig:acc_by_rank_L-14}), observing analogous findings. In fact, the first shows a decrease of 1.3\% mean accuracy at 3\% of retained TSVs and the second shows a reduction of 1.1\% mean accuracy at 2\% of maintained TSVs. We refer to \Cref{fig:rank_analysis_by_dataset} for a breakdown of this analysis.
\begin{figure}[htbp]
  \centering
  \includegraphics[width=\linewidth]{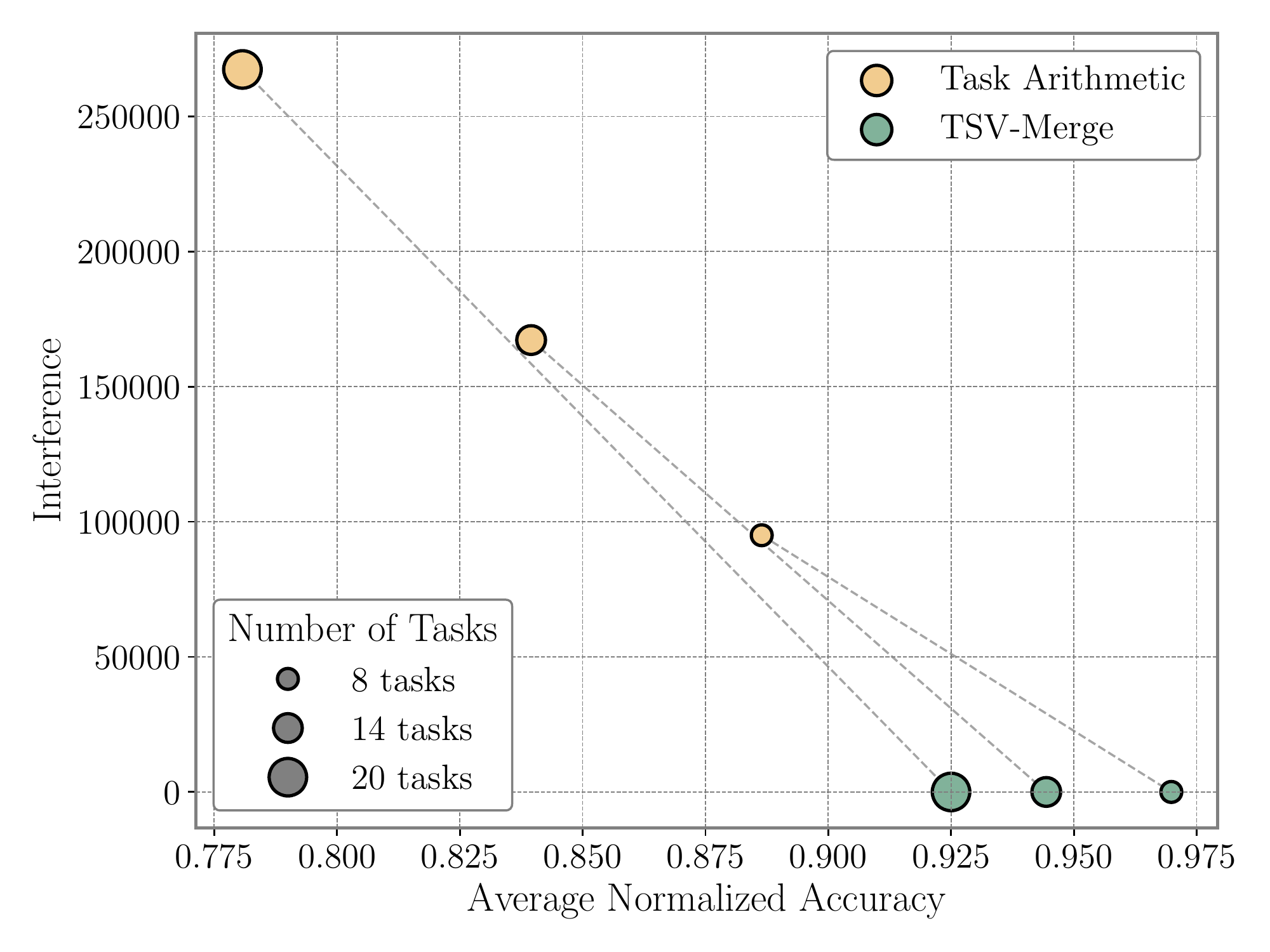}
  \caption{Singular Task Interference (STI) and average normalized accuracy for \baseline{Task Arithmetic} and \method{TSV-Merge} on the \model{ViT-L-14} model, evaluated across merges of 8, 14, and 20 tasks.}
  \label{fig:correlation_interference_accuracy_L-14}
\end{figure}
\subsubsection{Extended Ablation study}
In \Cref{subsec:ablation}, we reported an ablation study on the \model{ViT-B-32} model to evaluate the individual contributions of low-rank approximation and interference reduction to the overall performance of our \method{TSV-Merge} method. To further mark our findings and demonstrate the generality of our approach across different model sizes, we report in \Cref{tab:ablation-study-B-16} the ablation study for the \model{ViT-B-16} model and in \Cref{tab:ablation-study-L-14} \model{ViT-L-14} model. The experimental setup follows the one described in \Cref{subsec:ablation}. We assess the impact of the two key components of \method{TSV-M}, low-rank approximation and interference reduction, by considering the following four configurations:
\begin{enumerate}
    \item \textbf{Baseline Task Arithmetic}: the standard TA method without any modification.
    \item \textbf{Low-Rank Approximation}: apply only low-rank approximation to task matrices without any interference reduction step.
    \item \textbf{Interference Reduction}: apply interference reduction to the full-rank task matrices without any pre-step of low-rank approximation.
    \item \textbf{TSV-Merge}: Combining both low-rank approximation and interference reduction.
\end{enumerate}

\subsubsection{Effect of task interference}
We provide here the same plots shown in \Cref{fig:correlation_interference_accuracy}, for the \model{ViT-B-16} we show it in \Cref{fig:correlation_interference_accuracy_B-16} and respectively for \model{ViT-L-14} in \Cref{fig:correlation_interference_accuracy_L-14}. The finding remains valid also for these models, all the instances show a significant gain in accuracy when the interference is removed.

\subsubsection{Detailed per-layer task interference}
We show in \cref{fig:detailed_interference_by_layer} the per-layer task interference, extending the block-level analysis in \Cref{fig:interference_by_layer} in the main manuscript.

\subsubsection{Compression analysis}
\begin{figure}[t]
    \centering
    \includegraphics[width=\linewidth]{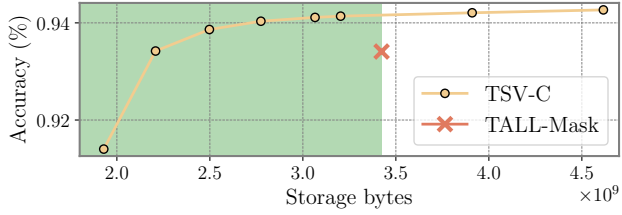}
    \vspace{-0.35cm}
    \caption{Accuracy with varying compression budgets for \model{ViT-L-14} across 14 tasks.}
    \label{fig:compression}
    \vspace{-0.35cm}
\end{figure}

Our experimental results (see \Cref{tab:task_acc_compression} in main manuscript) demonstrate that TSV outperforms TALL-Mask on the large-scale \model{ViT-L-14} model for 14 and 20 tasks benchmarks, signaling a scaling advantage. With a fixed-budget analysis, we show in \Cref{fig:compression} that unlike TALL-Mask, which has a fixed requirement defined by model size and number of tasks, we allow flexible compression rates by allowing rank selection. This enables more aggressive compression, as highlighted in the green region in the figure.

\subsubsection{Test-time adaptation}
We compare our method with \baseline{AdaMerging} \citep{yang2024adamergingadaptivemodelmerging} for test-time adaptation. On a subset of 7 tasks from the 8 task benchmark, \baseline{AdaMerging} achieves an accuracy of 85.43\%, while our \method{TSV-M} attains 88.93\%, an improvement of approximately 3.5\% without requiring any test-time adaptation. Additionally, when integrating an \baseline{AdaMerging}-style test-time adaptation into our framework, the accuracy increases to 89.87\%, demonstrating the complementary benefits of combining TSV-M with test-time adaptation techniques.

\renewcommand{\thesection}{E}
\section{Theoretical motivations and analysis}

\subsection{Theoretical foundation - Empirical design}
The \method{TSV-C} method is grounded in the well-established framework of low-rank approximation for compression (e.g., \cite{denton2014exploiting}). Instead, \method{TSV-M} is motivated by more empirical foundations: it is designed to achieve noise reduction through low-rank approximation and to eliminate interference via orthogonalization. Low-rank truncation serves to filter out insignificant variations, while orthogonalization ensures that task-specific singular vectors remain independent, preserving individual task performance.

\subsection{Heuristic interference measure}
Given that a formal definition of interference in model merging is not yet established, we adopt an operational definition: interference is any cross-task interaction that hinders the merging process. Our proposed Singular Task Interference measure is empirically validated by the consistent performance improvements observed when its value is minimized.
Furthermore, we examine the relationship between overlapping singular vectors and knowledge sharing. Unlike multi-task learning (MTL), which enables coordinated knowledge sharing through joint training, the independent task-wise finetuning in model merging may evolve in destructive overlaps in the activations, resulting in interference rather than beneficial knowledge sharing. By orthogonalizing the singular vectors, our approach effectively mitigates these overlaps, reducing interference and enhancing the performance of the merged model.

\begin{figure*}
  \centering
  \includegraphics[width=0.97\linewidth, height=0.99\textheight]{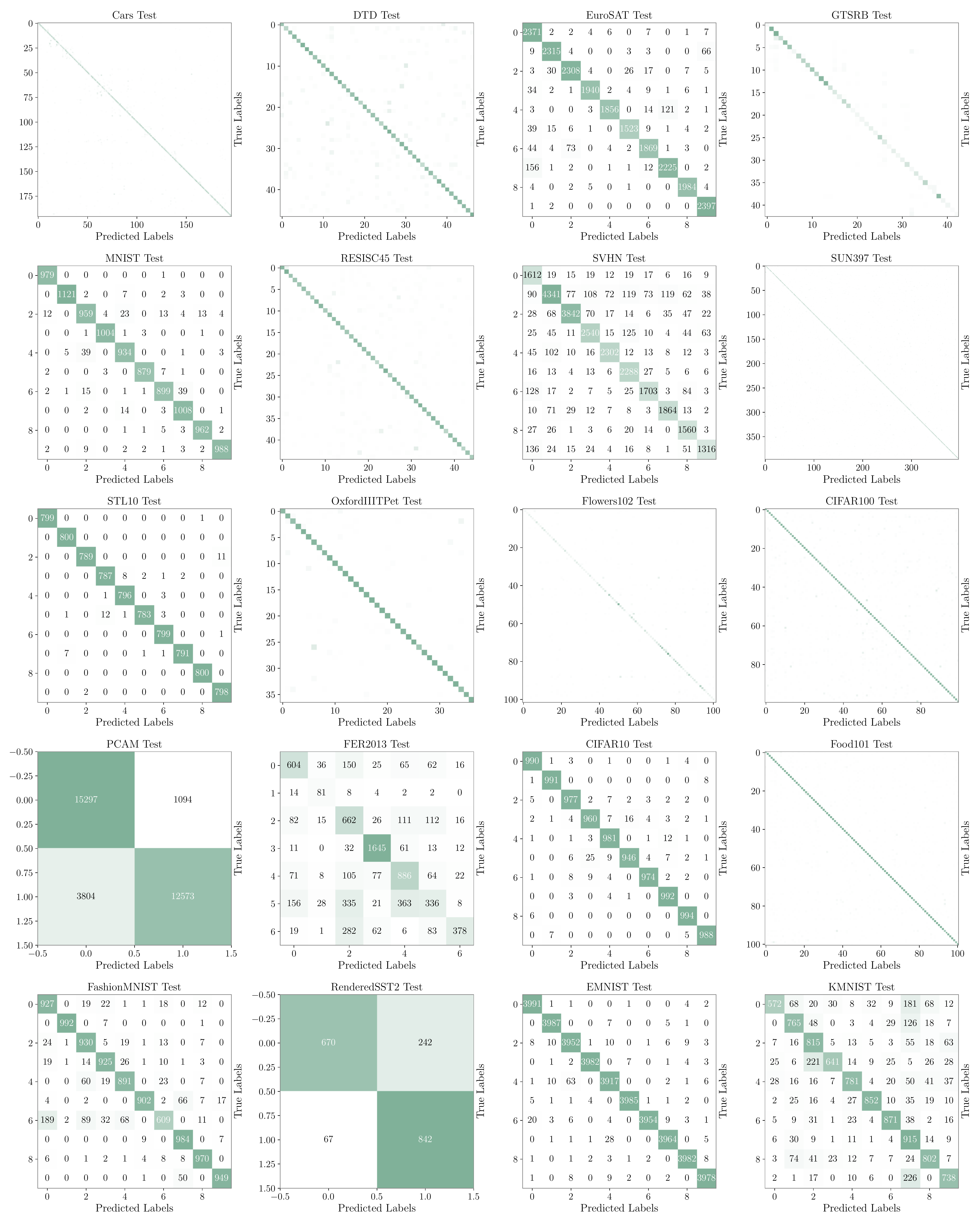}
  \caption{Breakdown of classification accuracy in confusion matrices of a single merged \model{ViT-L-14} model over 20 tasks. The numbers are omitted when they are too small to display.}
  \label{fig:confusion_matrix}
\end{figure*}

\end{document}